\documentclass[twocolumn,letterpaper]{article}

\usepackage{cvpr}              

\usepackage{colortbl}
\usepackage{xcolor}
\definecolor{cvprblue}{rgb}{0.21,0.49,0.74}
\usepackage[pagebackref,breaklinks,colorlinks,allcolors=cvprblue]{hyperref}

\def\paperID{****} 
\def\confName{CVPR}
\def\confYear{2026}

\title{Learning to Select Visual In-Context Demonstrations}

\author{
    Eugene Lee$^{1}$, Yu-Chi Lin$^{2}$, Jiajie Diao$^{1}$ \\
    $^{1}$ University of Cincinnati \quad
    $^{2}$ University of California, Los Angeles \\
    {\tt\small eugene.lee@uc.edu, yclin0177@g.ucla.edu, jiajie.diao@uc.edu}
    }

\newcommand\blfootnote[1]{%
  \begingroup
  \renewcommand\thefootnote{}\footnote{#1}%
  \addtocounter{footnote}{-1}%
  \endgroup
}

\begin{document}
\maketitle
\begin{abstract}
Multimodal Large Language Models (MLLMs) adapt to visual tasks via in-context learning (ICL), which relies heavily on demonstration quality. The dominant demonstration selection strategy is unsupervised k-Nearest Neighbor (kNN) search. While simple, this similarity-first approach is sub-optimal for complex factual regression tasks; it selects redundant examples that fail to capture the task's full output range. We reframe selection as a sequential decision-making problem and introduce Learning to Select Demonstrations (LSD), training a Reinforcement Learning agent to construct optimal demonstration sets. Using a Dueling DQN with a query-centric Transformer Decoder, our agent learns a policy that maximizes MLLM downstream performance. Evaluating across five visual regression benchmarks, we uncover a crucial dichotomy: while kNN remains optimal for subjective preference tasks, LSD significantly outperforms baselines on objective, factual regression tasks. By balancing visual relevance with diversity, LSD better defines regression boundaries, illuminating when learned selection is strictly necessary for visual ICL.
\end{abstract}    
\section{Introduction}
\blfootnote{Project page and code: \url{https://eugenelet.github.io/LSD-Project/}}
\label{sec:intro}


Multimodal Large Language Models (MLLMs) and Large Language Models (LLMs) have demonstrated remarkable abilities in complex tasks through in-context learning (ICL) \cite{dong2022survey}, including mathematical reasoning \cite{wei2022chain}. This paradigm has driven a significant shift in few-shot learning (FSL). With the advent of powerful Vision Foundation Models (VFMs) and Vision-Language Models (VLMs), ICL is now the dominant approach for few-shot adaptation. Consequently, as \cite{zhu2025exploring} notes, the core research question has pivoted from training few-shot learners to effectively prompting massively pre-trained models.

\begin{figure}[t!]
    \centering
    \includegraphics[width=\linewidth]{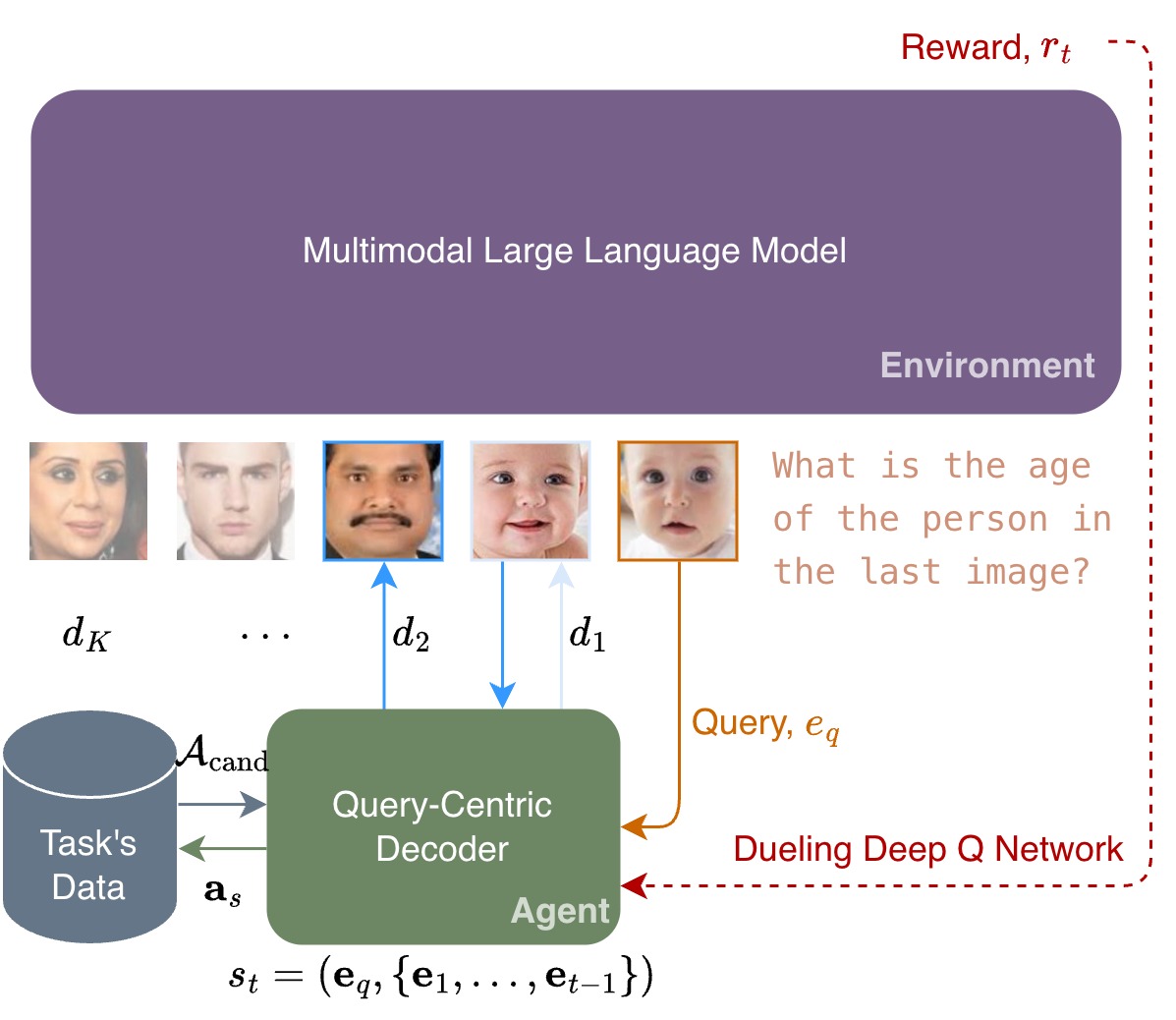}
    \caption{
    \textbf{An overview of our LSD (Learning to Select Demonstrations) framework.}
    The process is a training loop where the MLLM acts as the Environment.
    \textbf{(1)} The \textbf{Agent} (a Dueling DQN) receives the current state $s_t$, which contains the query embedding $\mathbf{e}_q$ and the embeddings of all previously selected demonstrations $\{\mathbf{e}_1, \dots, \mathbf{e}_{t-1}\}$.
    \textbf{(2)} The agent's query-centric decoder outputs an advantage query $\mathbf{a}_s$, which is used to retrieve candidates $A_{\text{cand}}$ from the \textbf{Task's Data} via FAISS.
    \textbf{(3)} The agent selects the next best demonstration, $d_t$.
    \textbf{(4)} The full prompt (including the selected demos $d_1 \dots d_K$ and the query) is sent to the \textbf{MLLM} (Environment), which makes a prediction.
    \textbf{(5)} A \textbf{Reward $r_t$} is calculated based on the prediction's accuracy (e.g., MAE).
    \textbf{(6)} This reward is used to update the agent's policy.
}
    \label{fig:overview}
\end{figure}

However, ICL efficacy is highly sensitive to prompt configuration, especially the selection and ordering of demonstration examples \cite{wang2023large,liu2024let}. The impact of effective ICL spans diverse applications, including data engineering \cite{wang2021want,khorashadizadeh2023exploring,ding2022gpt}, model augmentation \cite{ram2023context}, knowledge updating \cite{de2021editing}, model safety \cite{panda2023differentially,meade2023using}, and sentiment analysis \cite{zhang2023sentiment,xu2024improving,wang2024context,yang2024empirical}.

The most common selection strategy relies on unsupervised nearest neighbor (kNN) retrieval based on feature similarity \cite{liu2021makes,tanwar2023multilingual,qin2023context}. While simple, this approach is often sub-optimal due to a lack of task-specific supervision \cite{rubin2021learning,ye2023compositional,wang2023large,zhang2022active}. Its core ``similarity-priority'' assumption exhibits limited predictive power \cite{wu2025towards} and frequently yields redundant demonstration sets that provide misleading contextual information \cite{li2024context}.

To move beyond simple similarity, research has explored \emph{demonstration ordering}—arranging examples by proximity \cite{liu2021makes} or complexity \cite{liu2024let}—and \emph{demonstration construction}, emphasizing diversity \cite{an2023context} or using LLMs to generate new demonstrations \cite{kim2022self,yang2023auto,hao2022structured,yang2023representative}. For visual ICL, complex retrieval-reranking paradigms have been proposed \cite{zhou2024visual}, alongside metrics designed to select for ``representativeness'' \cite{guo2024makes} or to explicitly model structural complexity \cite{li2024context}.

A more fundamental critique, inspired by hard negative mining \cite{li2019deep}, argues these approaches over-index on positive, high-similarity examples. This has prompted a paradigm shift reframing shot selection as a sequential decision-making problem aimed at finding the most "informative" examples \cite{margatina2023active}. This view treats demonstration selection as a task for a Reinforcement Learning (RL) agent, learning a policy to maximize cumulative rewards tied to final ICL accuracy \cite{zhang2022active}, shifting retrieval from simple visual similarity to a more abstract, task-oriented ``reasoning process similarity'' \cite{qin2023context}.

We embrace this sequential paradigm to address a critical gap in visual ICL: understanding \emph{when} learned selection is actually necessary. Building on efforts utilizing LLM feedback \cite{zhang2022active, zhang2025learning} or RL frameworks \cite{wang2024demonstration}, we propose \emph{LSD (Learning to Select Demonstrations)}, a novel RL framework that trains a Dueling DQN agent to sequentially construct demonstration sets for visual regression tasks. Our key hypothesis is that the optimal selection strategy depends fundamentally on whether the task is \emph{objective} or \emph{subjective}. For objective, factual tasks, the optimal set must contain diverse ``boundary'' examples that help the MLLM model the entire regression space. Conversely, for subjective preference tasks, a simple visual anchor often suffices. As shown in Fig. \ref{fig:overview}, our agent uses a query-centric Transformer Decoder to learn a policy that actively balances visual relevance with necessary diversity, avoiding the redundancy trap of kNN to maximize accuracy on complex objective domains.

Our main contributions are:
\begin{itemize}
    \item We introduce LSD, a novel framework that successfully reframes $K$-shot demonstration selection as a sequential decision-making problem, scaling to dataset-level action spaces using a Dueling DQN agent and a query-centric Transformer Decoder.
    \item We conduct a comprehensive study on the efficacy of learned selection policies across five diverse visual regression benchmarks (UTKFace, AVA, SCUT-FBP5500, KonIQ-10k, and KADID-10k).
    \item We reveal a critical, task-dependent dichotomy in visual ICL: while unsupervised similarity search (kNN) remains highly effective for subjective preference tasks, our learned, diversity-aware policy is strictly necessary to achieve state-of-the-art performance on objective visual regression tasks.
\end{itemize}
\section{Related Work}
\label{sec:related}

Our research builds upon a large body of work in visual in-context learning (ICL), particularly on the critical problem of demonstration selection.

\paragraph{Demonstration Selection for In-Context Learning.}
The performance of ICL is known to be highly sensitive to the choice of demonstration examples \cite{zhang2023makes, liu2024let}. This has been shown in various domains, with recent work demonstrating that MLLMs like GPT-4V can classify specialized medical images (e.g., histopathology) with high accuracy using just a few well-chosen examples \cite{ferber2024context}. This sensitivity has spurred significant research into methods that move beyond simple kNN retrieval.

A primary challenge is selecting an optimal \emph{set} of demonstrations, not just individual relevant examples. One line of work treats this as a subset selection problem. Yang \etal \cite{yang2023representative} proposed selecting a single, representative set of demonstrations applicable to all test instances, using a Determinantal Point Process (DPP) to ensure both quality and diversity. This task-level approach contrasts with our instance-level policy. Purohit \etal \cite{purohitsample} (CASE) framed set selection as a multi-armed bandit problem, treating each subset as an ``arm'' and using a novel sampling strategy to efficiently find the best set while minimizing expensive LLM calls.

Another line of research explores the trade-off between the two main criteria for selection: similarity and diversity. While similarity-based retrieval is effective for simple tasks, Xiao \etal \cite{xiao2025role} systematically demonstrated that incorporating diversity is crucial for improving performance and robustness on complex tasks, such as math and code generation. This finding directly supports our hypothesis that a learned agent is necessary to intelligently balance these two competing objectives.

Rather than retrieving a static set, other methods treat selection as a sequential construction problem. Li \cite{liadvancing} introduced SabER, a lightweight decoder that autoregressively selects \emph{and} orders examples to construct an optimal prompt. While holistic, this approach is trained on scores from the target MLLM, making the resulting selector model-specific and requiring retraining for different MLLMs. Our RL-based approach, while also sequential, learns from a more generalizable reward signal (downstream MAE) and is not as tightly coupled to the reward model's architecture.

Finally, some work has focused on improving the retriever itself. Zhang \etal \cite{zhang2023makes} proposed a supervised, contrastive learning framework to train a retriever that automatically selects examples which maximize downstream task performance. This highlights the value of task-specific supervision, which our RL framework incorporates via its reward function, in contrast to unsupervised similarity metrics.

\paragraph{Related ICL Training and Prompting Strategies.}
Beyond demonstration selection, other methods aim to improve IDCL by modifying the model's training or the prompting method itself. To better leverage few-shot examples, Lin \etal \cite{doveh2024towards} introduced an ``any-shot'' training paradigm, showing that explicitly training models on ICL-formatted, multi-turn conversations enhances their ability to learn from context.

Diverging from selection entirely, other research explores how to elicit better reasoning from the LLM with no demonstrations. Yao \cite{yao2024large}, for instance, introduced Contrastive Prompting, a zero-shot method that instructs an LLM to generate both a correct and an incorrect solution. This process of explicit contrastive reasoning was shown to significantly boost performance on complex tasks by helping the model better discern the correct problem-solving path. Our work draws on a similar intuition: providing a diverse or ``contrastive'' set of demonstrations (e.g., high and low scores) in the prompt can serve a similar purpose, helping the model to ``triangulate'' the correct answer.
\section{Method}
\label{sec:method}

The problem of selecting an optimal set of $K$ demonstrations for in-context learning (ICL) can be framed as a sequential decision-making task. While unsupervised methods based on feature similarity are common \cite{liu2021makes,guo2024makes}, they are often sub-optimal as they lack task-specific supervision \cite{rubin2021learning,wang2023large}. To overcome this, we adopt a Reinforcement Learning (RL) framework, similar to approaches in \cite{zhang2022active,wang2024demonstration}, to learn a policy that iteratively constructs a high-quality demonstration set.

Our core contribution is a novel Dueling Deep Q-Network (DQN) \cite{wang2016dueling} architecture specifically designed to handle the massive, discrete action space inherent in demonstration selection, where any sample from the entire dataset $N$ can be chosen. Instead of a linear output layer of size $N$, our network computes Q-values by projecting the state representation into a query vector, which then interacts with the embedding of all possible actions via an efficient, approximate nearest-neighbor search.

\subsection{Problem Formulation as an MDP}
We model the $K$-shot demonstration selection process as a finite-horizon Markov Decision Process (MDP), defined by the tuple $(\mathcal{S}, \mathcal{A}, \mathcal{P}, \mathcal{R}, \gamma)$:

\begin{itemize}
    \item \textbf{State ($s_t \in \mathcal{S}$):} A state at step $t$ (for $t=1, \dots, K$) is defined by the query $q$ and the ordered set of demonstrations selected so far, $D_{t-1} = \{d_1, \dots, d_{t-1}\}$. The initial state $s_1$ contains the query and one ``anchor'' demonstration found via nearest-neighbor search, to provide initial context.

    \item \textbf{Action ($a_t \in \mathcal{A}$):} An action is the selection of a new demonstration $d_t$ from the pool of all available samples $\mathcal{C}$, excluding the query and any previously selected demonstrations: $a_t \in \mathcal{C} \setminus (\{q\} \cup D_{t-1})$. The action space $|\mathcal{A}|$ is thus $O(N)$, where $N$ is the total number of samples in the dataset.

    \item \textbf{Transition ($\mathcal{P}$):} The state transition is deterministic. Upon taking action $a_t = d_t$ in state $s_t = (q, D_{t-1})$, the environment transitions to state $s_{t+1} = (q, D_t)$, where $D_t = D_{t-1} \cup \{d_t\}$. The episode terminates when $K$ demonstrations have been selected ($t=K$).

    \item \textbf{Reward ($\mathcal{R}$):} The reward function is designed to optimize the marginal utility of each added demonstration. We define a MLLM scoring function, $R(s_t) = -\text{MAE}(\mathcal{V}(q, D_{t-1}))$, which queries the MLLM with the query $q$ and demonstrations $D_{t-1}$ and returns the negative Mean Absolute Error (MAE) of its prediction. The reward $r_t$ for selecting action $a_t$ is the \emph{improvement} in this score:
    \begin{equation}
        r(s_t, a_t) = R(s_{t+1}) - R(s_t)
        \label{eq:reward}
    \end{equation}
    This sparse reward encourages the agent to select samples that progressively refine the MLLM's accuracy. A large penalty is given for invalid actions (e.g., re-selecting a sample) or MLLM failures.

    \item \textbf{Discount ($\gamma$):} We use a discount factor $\gamma$ to balance immediate and future rewards.
\end{itemize}

The agent's goal is to learn the optimal action-value function $Q^*(s, a)$, which represents the maximum expected cumulative reward $G_t = \sum_{i=t}^K \gamma^{i-t} r_i$ from state $s$ by taking action $a$ and following the optimal policy thereafter.

\subsection{Dueling Q-Network for Large Action Spaces}
A standard DQN is infeasible due to the $O(N)$ action space. We therefore employ a Dueling Q-Network architecture that leverages the underlying embedding space $\mathbb{R}^D$ of the samples. All $N$ samples are represented by a $D$-dimensional embedding $\mathbf{e}_i$, pre-computed using a SigLIP model \cite{zhai2023sigmoid}.

The $Q(s, a)$ function is decomposed into a state-value $V(s)$ and an action-advantage $A(s, a)$ \cite{wang2016dueling}:
\begin{equation}
    Q(s, a) = V(s) + \left( A(s, a) - \frac{1}{|\mathcal{A}|} \sum_{a' \in \mathcal{A}} A(s, a') \right)
    \label{eq:dueling}
\end{equation}

Our network (\cref{fig:overview}) does not compute $A(s, a)$ for all $a \in \mathcal{A}$. Instead, it computes $V(s)$ and a $D$-dimensional ``advantage query'' vector $\mathbf{a}_s$. The advantage $A(s, a_i)$ for a specific action (sample $i$) is then calculated as the inner product of this query with the action's embedding $\mathbf{e}_i$:
\begin{equation}
    A(s, a_i) = \mathbf{a}_s^\top \mathbf{e}_i
    \label{eq:advantage_dot}
\end{equation}
This formulation assumes both $\mathbf{a}_s$ and $\mathbf{e}_i$ are L2-normalized, making the advantage a measure of cosine similarity.

\subsection{Network Architecture}
Our network consists of a query-centric state encoder and two dueling heads.

\subsubsection{Query-Centric State Encoder}
\label{sec:state_encoder}

To produce a holistic state representation, we must fuse the query embedding $\mathbf{e}_q$ with the set of $t-1$ selected demonstration embeddings $\mathbf{E}_D = \{\mathbf{e}_1, \dots, \mathbf{e}_{t-1}\}$.

A common approach might be to simply concatenate these embeddings, $[\mathbf{e}_q; \mathbf{e}_1; \dots; \mathbf{e}_{t-1}]$, and pass them through a Transformer Encoder (i.e., using only self-attention). However, our initial experiments revealed a significant failure mode with this design: the agent was prone to \emph{policy collapse}, learning to select a single, query-agnostic set of "generally good" demonstrations, regardless of the query's specific features. This indicates the self-attention mechanism failed to adequately prioritize the query's relationship to the demonstrations.

To solve this, we designed a \emph{Query-Centric State Encoder} using a standard Transformer Decoder architecture \cite{vaswani2017attention} in a specific way. We feed the L2-normalized query embedding $\mathbf{e}_q$ as the \emph{target} sequence (with a sequence length of 1) and the set of $t-1$ demonstration embeddings $\mathbf{E}_D$ as the \emph{memory} sequence. As ICL is sensitive to demonstration order \cite{liu2021makes,liu2024let}, the demonstration embeddings are first augmented with a learned positional encoding $\mathbf{P} \in \mathbb{R}^{K \times D}$.

A standard Transformer Decoder layer contains three sub-layers: masked self-attention, cross-attention, and a feedforward network (FFN). Because our \emph{target} sequence has a length of one, the initial \emph{masked self-attention sub-layer is definitionally bypassed} (it's a no-op).

Therefore, the computation within each of the $L$ decoder layers reduces to two critical steps:
\begin{enumerate}
    \item \emph{Cross-Attention:} The query representation $\mathbf{x}_q^{(l-1)}$ (from the previous layer) is used to generate the \emph{Query (Q)} vector. This $\mathbf{Q}$ probes the \emph{memory} (demos) $\mathbf{M} = \mathbf{E}_D + \mathbf{P}$, which provides the \emph{Key (K)} and \emph{Value (V)} vectors. This step computes an attention-weighted vector that represents the query contextualized by the demonstrations.
    \item \emph{Feedforward Network (FFN):} The resulting vector is then processed by a standard position-wise FFN to produce the layer's output, $\mathbf{x}_q^{(l)}$.
\end{enumerate}
This process repeats for $L$ layers, progressively refining the query embedding based on the provided demonstration context. The final output $\mathbf{c}_s$ is the fully contextualized query vector, which is then passed to the dueling heads. This computation, performed by the $L$-layer $\text{TransformerDecoder}$, is defined as:
\begin{equation}
\label{eq:state_encoder}
\begin{aligned}
    \mathbf{c}_s &= \text{TransformerDecoder}(\text{target}=\mathbf{e}_q, \text{memory}=\mathbf{M}) \\
    \mathbf{M} &= \mathbf{E}_D + \mathbf{P} \, ; \quad \mathbf{x}_q^{(0)} = \mathbf{e}_q \\
    \mathbf{x}'_q &= \mathbf{x}_q^{(l-1)} + \text{Softmax}\left(\frac{(\mathbf{x}_q^{(l-1)}W_Q^{(l)}) (\mathbf{M}W_K^{(l)})^\top}{\sqrt{d_k}}\right) (\mathbf{M}W_V^{(l)}) \\
    \mathbf{x}_q^{(l)} &= \mathbf{x}'_q + \text{FFN}^{(l)}(\mathbf{x}'_q) \quad \forall l \in \{1, \dots, L\} \\
    \mathbf{c}_s &= \mathbf{x}_q^{(L)}
\end{aligned}
\end{equation}
where $\text{FFN}^{(l)}$ is the feedforward network for layer $l$. This design ensures the state representation $\mathbf{c}_s$ is always conditioned on the specific query, mitigating policy collapse and enabling the agent to learn a query-specific selection policy.

\subsubsection{Dueling Heads}
The context vector $\mathbf{c}_s$ is passed to two separate heads:
\begin{enumerate}
    \item \textbf{Value Head:} A simple linear layer that estimates the state-value $V(s)$:
    \begin{equation}
        V(s) = \mathbf{w}_v^\top \mathbf{c}_s + b_v
    \end{equation}
    where $\mathbf{w}_v \in \mathbb{R}^D$ and $b_v$ is a scalar bias.

    \item \textbf{Advantage Head:} A linear layer followed by L2 normalization, which produces the $D$-dimensional advantage query vector $\mathbf{a}_s$:
    \begin{equation}
        \mathbf{a}_s = \frac{\mathbf{W}_a \mathbf{c}_s + \mathbf{b}_a}{\|\mathbf{W}_a \mathbf{c}_s + \mathbf{b}_a\|_2}
    \end{equation}
    where $\mathbf{W}_a \in \mathbb{R}^{D \times D}$ and $\mathbf{b}_a \in \mathbb{R}^D$.
\end{enumerate}

\subsection{Approximate Q-Learning for Large Action Spaces}
\label{sec:training_approx}

The Q-learning update requires computing the target value $y_t$, which depends on $\max_{a'} Q(s', a')$. Finding the true maximum would require $N$ dot products (\cref{eq:advantage_dot}), which is computationally prohibitive.

To solve this, we leverage Approximate Nearest Neighbor (ANN) search. We build a FAISS (IVFPQ) index \cite{douze2024faiss} on the embeddings $\{\mathbf{e}_i\}_{i=1}^N$ of all dataset samples. This index can efficiently retrieve the $\mathcal{N}$ candidate actions whose embeddings have the highest inner product with a given advantage query vector $\mathbf{a}_s$.

\subsubsection{Action Selection}
We use an $\epsilon$-greedy policy. With probability $\epsilon$, we explore by selecting a random valid action from the $\mathcal{N}$ candidates returned by FAISS. With probability $1-\epsilon$, we exploit by executing the following steps:
\begin{enumerate}
    \item Compute the state-value $V(s_t)$ and the advantage query $\mathbf{a}_{s_t}$ using the policy network $Q_\theta$.
    \item Use the FAISS index to retrieve the top $\mathcal{N}$ candidate actions: \newline $\mathcal{A}_{\text{cand}} = \text{FAISS}(\mathbf{a}_{s_t}, \mathcal{N})$.
    \item Calculate the advantage $A(s_t, a_j)$ for all $a_j \in \mathcal{A}_{\text{cand}}$ using \cref{eq:advantage_dot}.
    \item Approximate the mean advantage using only these candidates: \newline $\bar{A} \approx \frac{1}{\mathcal{N}} \sum_{a_j \in \mathcal{A}_{\text{cand}}} A(s_t, a_j)$.
    \item Select the best action $a_t$ according to the dueling Q-value (\cref{eq:dueling}): 
    $$
    a_t = \text{argmax}_{a_j \in \mathcal{A}_{\text{cand}}} \left( V(s_t) + (A(s_t, a_j) - \bar{A}) \right)
    $$
\end{enumerate}

\subsubsection{Optimization}
We store transitions $(s_t, a_t, r_t, s_{t+1}, \text{done})$ in a replay buffer $\mathcal{B}$. For a mini-batch of $B$ transitions, we compute the target $y_t$ using the target network $Q_{\theta^-}$:
\begin{equation}
    y_t = r_t + \gamma (1 - \text{done}) \cdot \max_{a' \in \mathcal{A}'_{\text{cand}}} Q(s_{t+1}, a'; \theta^-)
    \label{eq:bellman_target}
\end{equation}
where the $\max$ operation is performed efficiently using the same FAISS-based approximation on the target network's advantage query $\mathbf{a}_{s'}$.

The policy network $Q_\theta$ is then updated by minimizing the Smooth L1 (Huber) Loss between the predicted $Q(s_t, a_t; \theta)$ and the target $y_t$:
\begin{equation}
    L(\theta) = \frac{1}{B} \sum_{(s,a,r,s') \in \mathcal{B}} \mathcal{L}_{\text{Huber}} \left( y_t - Q(s_t, a_t; \theta) \right)
    \label{eq:loss}
\end{equation}
The target network weights $\theta^-$ are updated via a soft polyak average: $\theta^- \leftarrow \tau \theta + (1 - \tau) \theta^-$.

\section{Experiments}
\label{sec:exp}

\begin{table*}[t!]
    \centering
    \caption{
        \textbf{Main Performance (MAE $\downarrow$) Comparison vs. Number of Shots ($K$).}
        We report the Mean Absolute Error (MAE) for all methods on the five benchmark datasets, evaluated with Gemma 3 4B-it for $K \in \{1, 4, 8, 16\}$. Our proposed method, \textbf{LSD}, consistently outperforms all baselines, and the performance gap widens as $K$ increases. The 0-shot and a fully \textbf{Supervised} (Sup.) baseline are also provided for reference. Best results are in \textbf{bold}.
    }
    \label{tab:main_results}
    \resizebox{\textwidth}{!}{%
    \begin{tabular}{l|c|c|cccc|cccc|cccc}
        \toprule
        & & & \multicolumn{4}{c|}{\textbf{Random}} & \multicolumn{4}{c|}{\textbf{kNN}} & \multicolumn{4}{c}{\textbf{LSD (Ours)}} \\
        \cmidrule(lr){4-7} \cmidrule(lr){8-11} \cmidrule(lr){12-15}
        \textbf{Dataset} & \textbf{Sup.} & \textbf{0-Shot}
            & K=1 & K=4 & K=8 & K=16
            & K=1 & K=4 & K=8 & K=16
            & K=1 & K=4 & K=8 & K=16 \\
        \midrule
        UTKFace & 4.42 \cite{paplham2024call} & 6.10
            & 6.14 & 10.66 & 14.86 & 12.51
            & 5.98 & 7.27 & 7.61 & 7.60
            & \textbf{5.90} & \textbf{6.27} & \textbf{7.05} & \textbf{6.64} \\
        AVA & - & 1.32
            & 1.21 & 1.03 & 0.92 & 0.92
            & \textbf{1.20} & \textbf{0.98} & \textbf{0.83} & \textbf{0.86}
            & \textbf{1.20} & 1.06 & 0.98 & 0.98 \\
        SCUT-FBP5500 & 0.26 \cite{xu2018transferring} & 0.59
            & 0.58 & 0.64 & 0.64 & 0.68
            & \textbf{0.53} & \textbf{0.39} & \textbf{0.40} & \textbf{0.44}
            & 0.55 & 0.62 & 0.67 & 0.75 \\
        KonIQ-10k & 0.39 \cite{tu2021regression} & 0.42
            & 0.42 & 0.48 & \textbf{0.44} & 0.56
            & 0.40 & 0.44 & 0.55 & 0.61
            & \textbf{0.39} & \textbf{0.40} & 0.51 & \textbf{0.51} \\
        KADID-10k & - & 0.94
            & 0.89 & 1.07 & 1.05 & 1.07
            & 0.87 & 0.87 & 0.91 & 0.92
            & \textbf{0.76} & \textbf{0.79} & \textbf{0.82} & \textbf{0.84} \\
        \bottomrule
    \end{tabular}
    }
\end{table*}

\begin{figure*}[t!]
    \centering
    \begin{subfigure}[b]{0.24\linewidth}
        \centering
        \includegraphics[width=\linewidth]{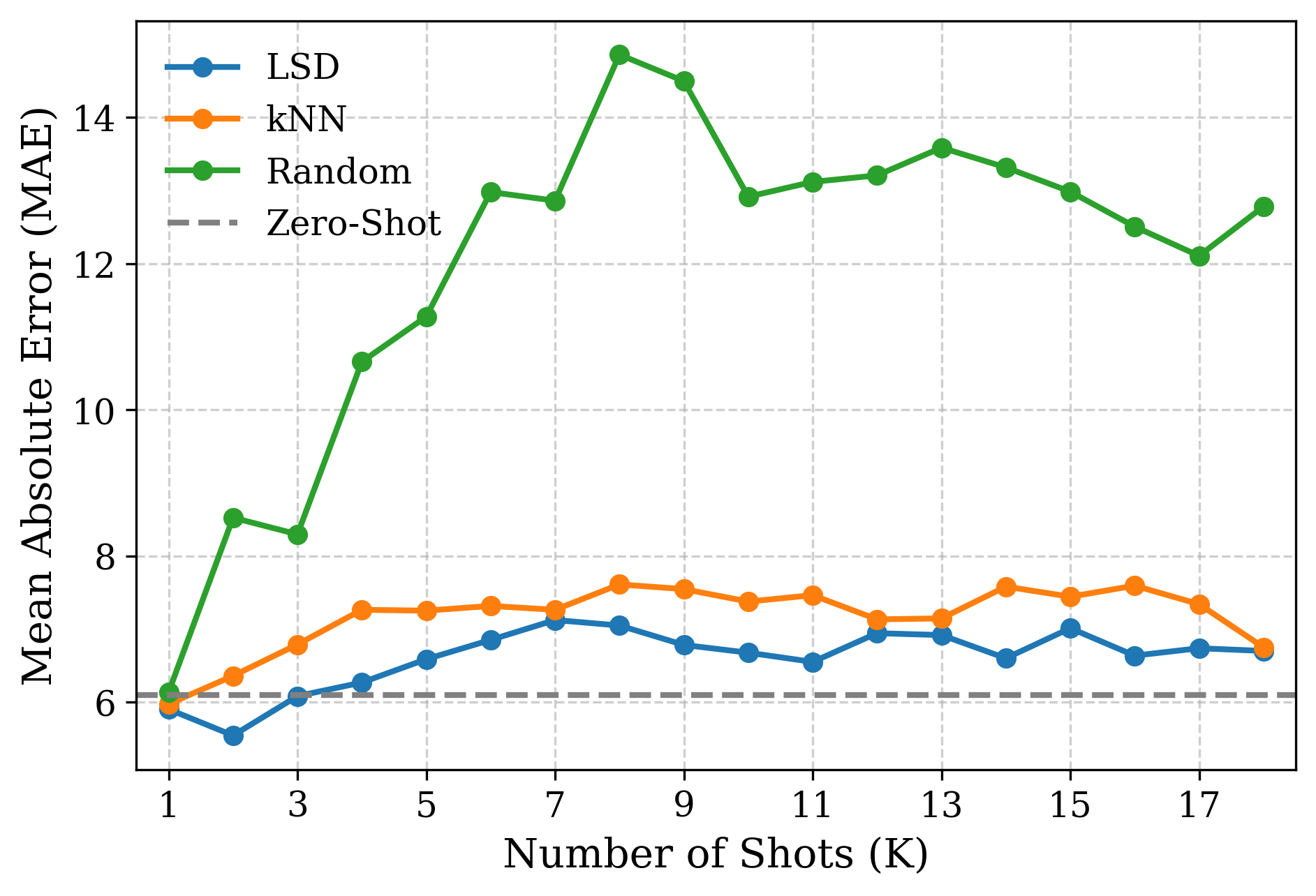}
        \caption{UTKFace (Age Prediction)}
        \label{fig:scaling_utk}
    \end{subfigure}
    \begin{subfigure}[b]{0.24\linewidth}
        \centering
        \includegraphics[width=\linewidth]{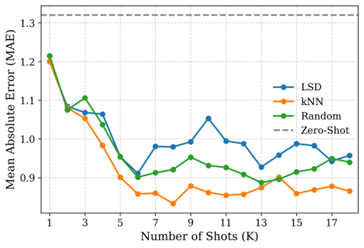}
        \caption{AVA (Aesthetic Rating)}
        \label{fig:scaling_ava}
    \end{subfigure}
    \begin{subfigure}[b]{0.24\linewidth}
        \centering
        \includegraphics[width=\linewidth]{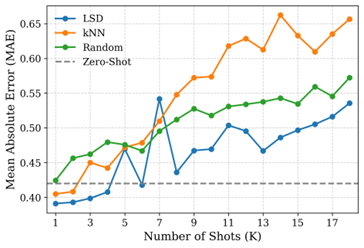}
        \caption{KonIQ-10k (Image Quality)}
        \label{fig:scaling_koniq}
    \end{subfigure}
    \begin{subfigure}[b]{0.24\linewidth}
        \centering
        \includegraphics[width=\linewidth]{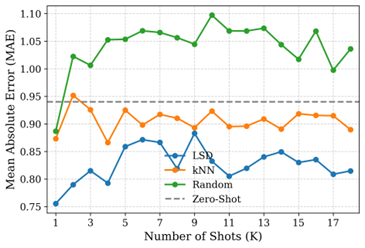}
        \caption{KADID-10k (Image Quality)}
        \label{fig:scaling_kadid}
    \end{subfigure}
    
    \caption{
        \textbf{Performance vs. Number of Shots ($K$) on four datasets.}
        We plot the MAE as $K$ increases. The results are task-dependent:
        \textbf{(a), (c), (d) Objective Tasks (UTKFace, KonIQ, KADID):} Our LSD policy (blue) consistently outperforms the kNN baseline (orange).
        \textbf{(b) Subjective Task (AVA):} The kNN baseline, which is based on visual similarity, consistently outperforms LSD.
    }
    \label{fig:scaling}
\end{figure*}

We conduct a comprehensive set of experiments to evaluate the effectiveness of our proposed demonstration selection method, which we refer to as \emph{LSD} (Learning to Select Demonstrations). Our evaluation is designed to answer several key questions:
\begin{enumerate}
    \item Does our method outperform standard unsupervised (kNN) and random selection baselines in terms of downstream task performance?
    \item How does the performance scale with the number of demonstrations ($K$)?
    \item Does our agent learn a selection policy that is qualitatively different from the baselines (e.g., by balancing relevance and diversity)?
    \item Can a policy learned using reward signals from one MLLM generalize to improve the performance of other, unseen MLLMs?
\end{enumerate}
To answer these, we evaluate on a diverse set of challenging visual regression tasks and compare against strong baselines.

\subsection{Datasets}
We focus on visual regression tasks, as they require nuanced reasoning from the MLLM that is often highly sensitive to demonstration quality. We use five public benchmark datasets:

\begin{itemize}
    \item \textbf{UTKFace:} A large-scale face dataset with over 20,000 images, annotated with age, gender, and ethnicity. For our experiments, we use the \emph{age prediction} task, which features a wide regression range from 0 to 116 years \cite{zhifei2017cvpr}.

    \item \textbf{AVA (Aesthetic Visual Analysis):} A large-scale database of over 250,000 images, annotated with aesthetic scores (a regression task on a 1-10 scale), as well as semantic labels and photographic styles \cite{murray2012ava}.

    \item \textbf{SCUT-FBP5500:} A facial beauty perception dataset consisting of 5,500 images of both Asian and Caucasian faces. Each image is annotated with an \emph{attractiveness rating} on a 1-to-5 scale, providing a fine-grained regression task \cite{liang2018scut}.

    \item \textbf{KonIQ-10k \& KADID-10k:} Two large-scale Image Quality Assessment (IQA) datasets. KonIQ-10k contains 10,073 images with quality scores obtained via crowdsourcing, reflecting ``authentic'' perceptual quality \cite{hosu2020koniq}. KADID-10k contains 10,000 images generated from 81 pristine images, each distorted by 25 different degradation types at 5 levels, providing a benchmark for "synthetic" distortion \cite{lin2019kadid}.
\end{itemize}

\subsection{Baselines}
We compare the performance of our method, \textbf{LSD}, against two standard and widely-used demonstration selection baselines:

\begin{itemize}
    \item \textbf{k-Nearest Neighbors (kNN):} This is the most common unsupervised baseline, based on the method in \cite{liu2021makes}. For a given query, we compute its SigLIP embedding and select the $K$ samples from the training pool with the highest cosine similarity to the query embedding.

    \item \textbf{Random:} We randomly select $K$ demonstrations from the training pool, excluding the query itself. This baseline helps establish whether the task benefits from ICL at all.

    \item \textbf{0-Shot:} We also report the performance of the MLLM with no demonstrations, which serves as the absolute performance floor and quantifies the overall benefit of ICL.
\end{itemize}

\subsection{Implementation Details}
\label{sec:impl_details}

\noindent\textbf{MLLMs.} Our experiments utilize three publicly available Multimodal Large Language Models: \emph{Gemma 3 4B-it} \cite{team2025gemma}, \emph{Qwen 2.5 7B} \cite{bai2025qwen2}, and \emph{Phi-3.5-vision (4.2B)} \cite{abdin2024phi}. Unless otherwise specified, our LSD agent is trained using reward signals generated by Gemma 3 4B-it, as described in \cref{sec:method}.

\noindent\textbf{Embeddings.} All sample embeddings for both our method and the kNN baseline are $D=768$ dimensional vectors extracted from the \emph{SigLIP-base-patch16-224} vision model.

\noindent\textbf{LSD Agent.} Our Dueling DQN agent's state encoder is a Transformer Decoder with $L=2$ layers and $H=4$ attention heads. We use a discount factor $\gamma = 0.99$, a learning rate of $5 \times 10^{-6}$, a replay buffer of 50,000 transitions, and a batch size of 32. For efficient action selection (\cref{eq:bellman_target}), we use a FAISS (IVFPQ) index to retrieve $\mathcal{N}=200$ candidates at each step. The agent is trained for 16,000 steps, which takes approximately 7 hours on a single NVIDIA A100 GPU.

\subsection{Evaluation Protocols}

We design several experiments to rigorously evaluate our method. While benchmarks like AVA, SCUT-FB5500, KonIQ-10k, and KADID-10k often report the Pearson Linear Correlation Coefficient (PLCC) or Spearman Rank Correlation Coefficient (SRCC), these metrics primarily measure the \emph{monotonicity} or \emph{correlation} of predictions against ground truth labels.

For our experiments, the primary metric is \emph{Mean Absolute Error (MAE)}. This choice is critical as it directly aligns with our method's optimization objective. Our RL agent's reward signal is a function of the MLLM's prediction error (e.g., $r_t \propto -\text{MAE}$). Therefore, evaluating with MAE is the most direct and accurate measure of our agent's success, as it quantifies exactly what the policy was trained to improve: the absolute accuracy of the MLLM's prediction.

\subsubsection{Main Performance vs. $K$}
Our primary experiment evaluates MAE as a function of the number of demonstrations $K$. The full results are presented in \cref{tab:main_results}, with performance scaling on key datasets shown in \cref{fig:scaling}. The results reveal a clear, task-dependent pattern.

On the \emph{objective} regression tasks—age prediction (UTKFace), and image quality assessment (KonIQ-10k and KADID-10k)—our learned LSD policy consistently and significantly outperforms the kNN baseline. As shown in \cref{tab:main_results}, this performance gap is evident across $K=4, 8, \text{and } 16$ for UTKFace and across all $K$ values for the IQA tasks. This highlights the efficiency of our learned, diversity-aware policy for tasks with a clear, factual ground truth.

Conversely, on the \emph{subjective} tasks that rely on human judgment, such as aesthetic rating (AVA) and attractiveness rating (SCUT-FBP5500), the kNN baseline provides superior performance. This suggests that for these tasks, simple visual similarity (which kNN excels at) is a more effective strategy than our agent's learned policy.

\begin{figure}[t!]
    \centering
    \begin{subfigure}[b]{0.48\linewidth}
        \centering
        \includegraphics[width=\linewidth]{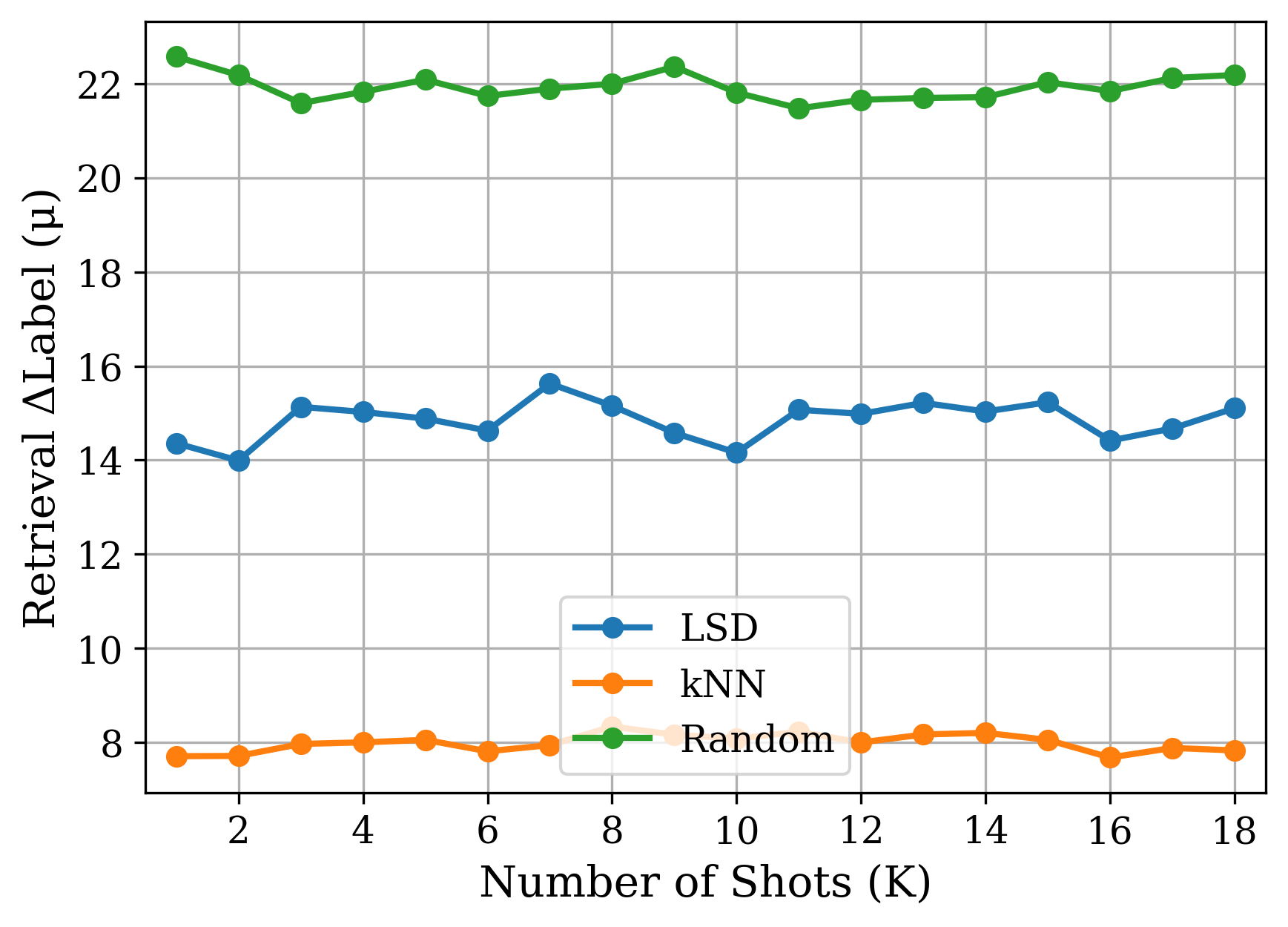}
        \caption{MAE of Demo Labels vs.\ Query}
        \label{fig:set_analysis_a}
    \end{subfigure}
    \hfill 
    \begin{subfigure}[b]{0.48\linewidth}
        \centering
        \includegraphics[width=\linewidth]{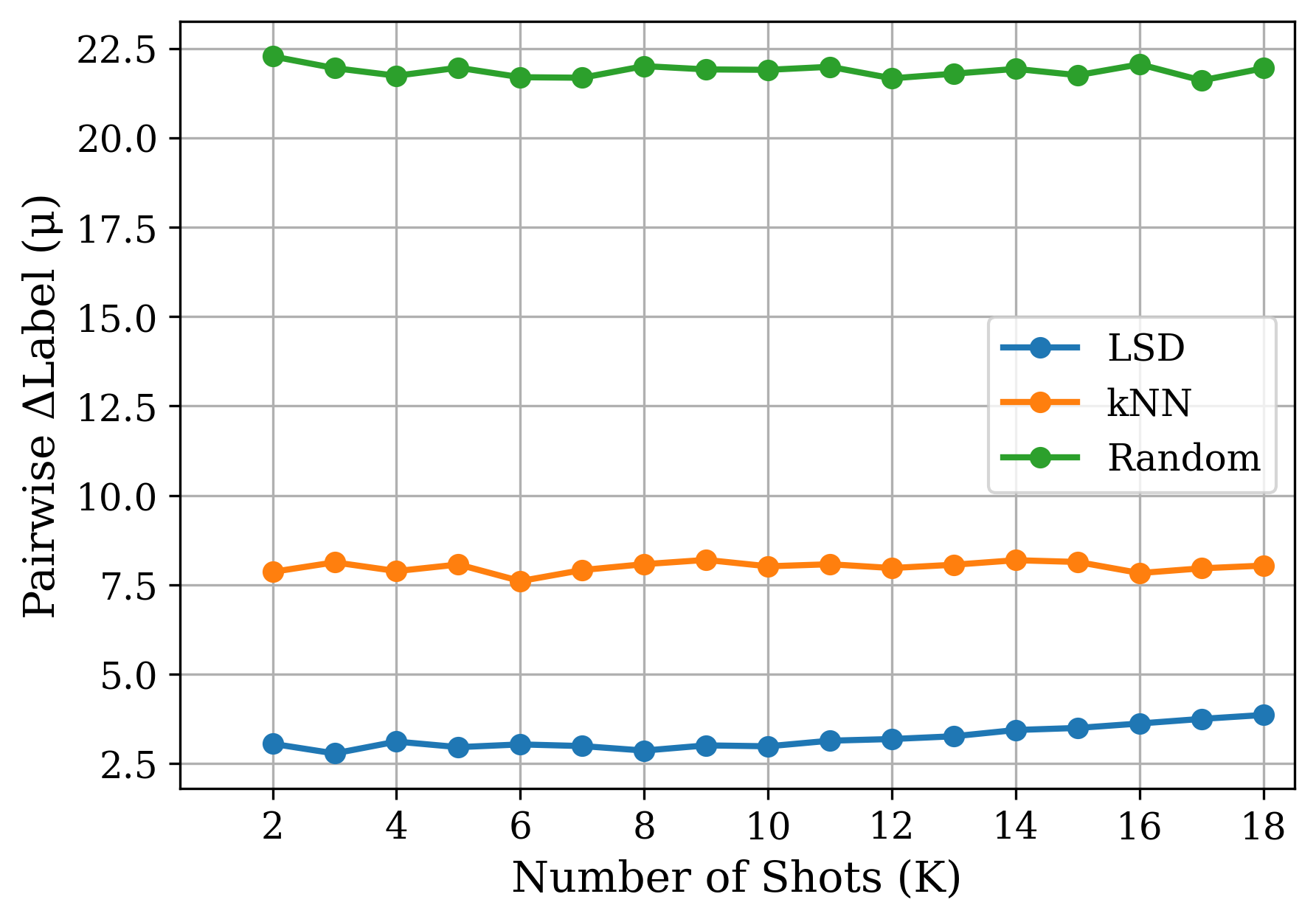}
        \caption{Pairwise Label MAE}
        \label{fig:set_analysis_b}
    \end{subfigure}
    
    \vspace{1em} 
    
    \begin{subfigure}[b]{0.48\linewidth}
        \centering
        \includegraphics[width=\linewidth]{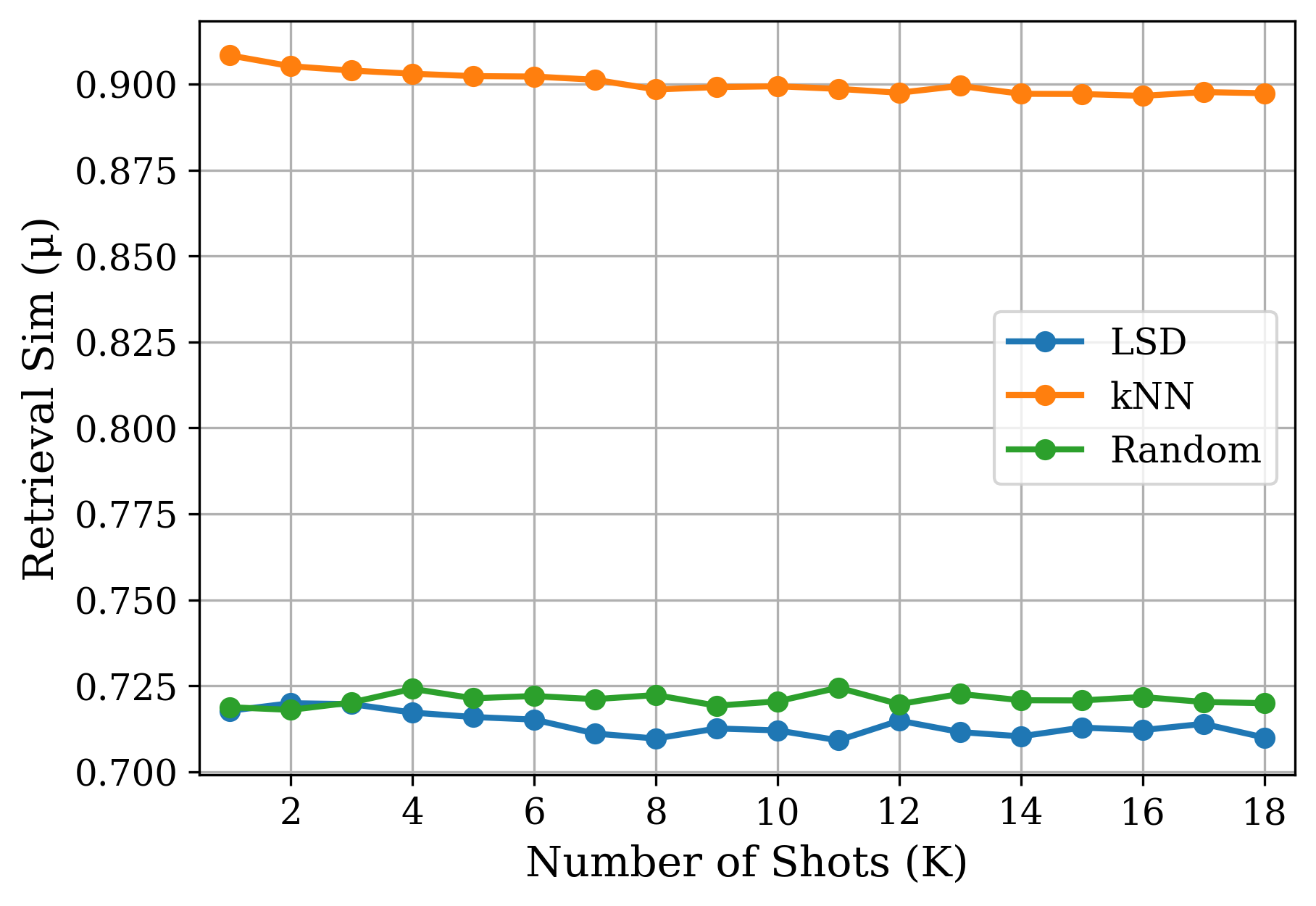}
        \caption{Demo-Query Feature Similarity}
        \label{fig:set_analysis_c}
    \end{subfigure}
    \hfill 
    \begin{subfigure}[b]{0.48\linewidth}
        \centering
        \includegraphics[width=\linewidth]{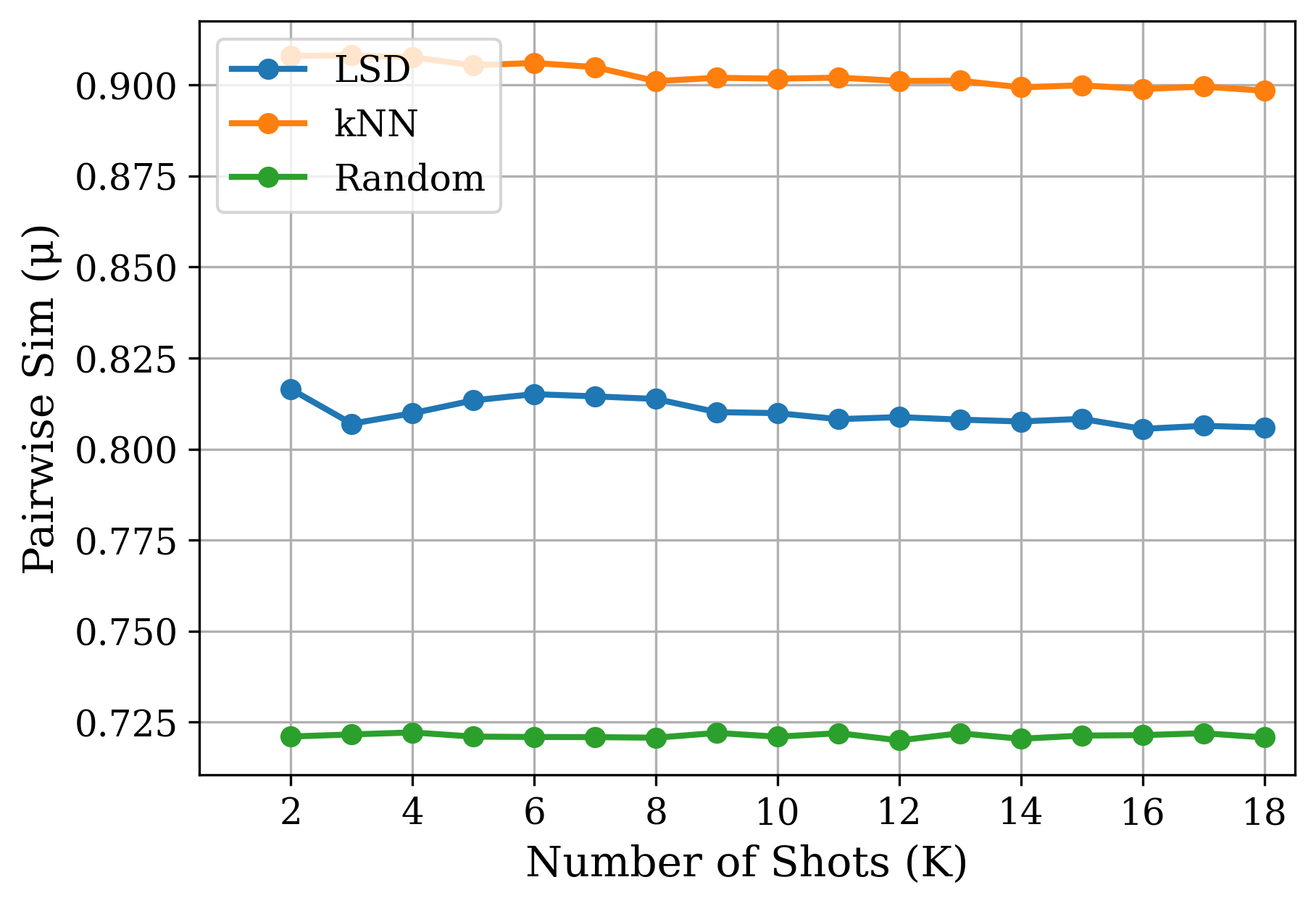}
        \caption{Pairwise Feature Similarity}
        \label{fig:set_analysis_d}
    \end{subfigure}
    
    \caption{
        \textbf{Demonstration Set Analysis on UTKFace, plotted against $K$ shots.}
        (a) \emph{MAE of Demo Labels vs.\ Query:} The MAE between selected demo labels and the query's true label. LSD finds demos with closer labels.
        (b) \emph{Pairwise Label MAE:} The MAE computed over all pairwise label differences among the selected demos.
        (c) \emph{Demo-Query Feature Similarity:} The cosine similarity between demo embeddings and the query embedding. LSD balances similarity with other factors.
        (d) \emph{Pairwise Feature Similarity:} The cosine similarity between every pair of selected demonstrations. LSD actively seeks diverse (low-similarity) demos.
    }
    \label{fig:set_analysis}
\end{figure}
\subsubsection{Demonstration Set Analysis}
To understand our agent's policy, we analyzed the selected demo sets on UTKFace (\cref{fig:set_analysis}). Our analysis shows kNN uses a fixed, myopic strategy (visual similarity), while LSD learns a sophisticated policy by optimizing for MLLM performance.

\begin{itemize}
    \item \textbf{Visual Relevance vs. Diversity:} kNN selects highly redundant, visually similar demos. In contrast, LSD learns a more nuanced policy: it prioritizes relevance (\cref{fig:set_analysis}(c)) but also actively seeks \emph{diversity} by selecting new demos that are visually \emph{dissimilar} from those already in the context (\cref{fig:set_analysis}(d)).
    \item \textbf{Emergent Label-Awareness:} Most strikingly, \cref{fig:set_analysis}(a) shows that by optimizing for the final reward, LSD \emph{implicitly learns} to select demos that are closer in label-space to the query (lower MAE), despite its state containing no label information.
\end{itemize}

In short, LSD learns a superior policy that balances visual relevance with active diversity, resulting in an emergent strategy highly correlated with the task's underlying label structure.
\begin{figure*}[]
    \centering
    \includegraphics[width=.8\linewidth]{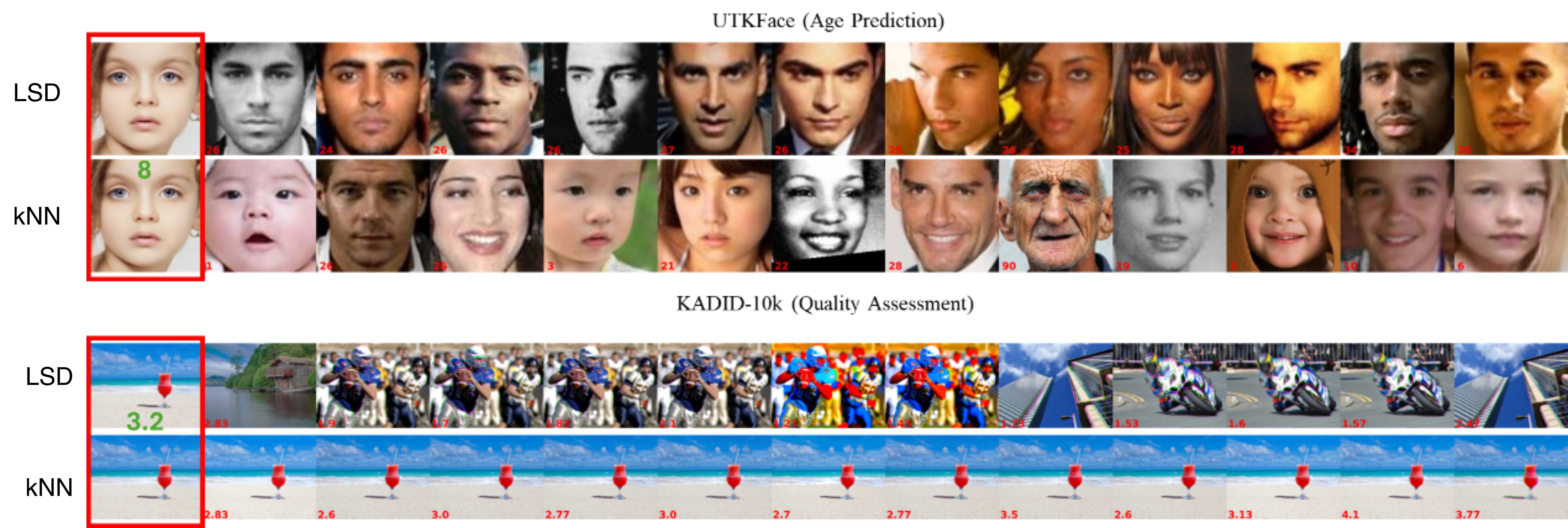}
    \caption{
        \textbf{Qualitative Comparison of Selected Demonstrations ($K=12$).}
        (a) \textbf{UTKFace:} For an 8-year-old query, kNN selects only images with highly similar features (e.g., other young children). LSD selects a diverse spectrum of visual features (e.g., varied ages, genders, and lighting conditions) to build a richer context.
        (b) \textbf{KADID-10k:} For a motion-blurred query, kNN selects only other distorted versions of the \emph{same source image}. LSD selects a varied set, including the pristine original and images with \emph{different distortion types} from \emph{different source images}, defining the quality boundaries.
    }
    \label{fig:qualitative}
\end{figure*}

\subsubsection{Qualitative Analysis}
Qualitative examples in \cref{fig:qualitative} illustrate the core policy differences. The \emph{kNN} baseline is myopic, invariably selecting a visually homogeneous and redundant set. For the 8-year-old query, it selects only other visually similar children. For the KADID-10k query, its policy is even more redundant, selecting only other distorted versions of the \emph{same source image}.

In contrast, our \emph{LSD} agent learns a sophisticated, context-building policy. For the UTKFace query, it selects a diverse spectrum of visual features, providing varied ages and appearances to help the MLLM understand the concept of ``age''. For the KADID-10k query, it learns to select crucial ``boundary'' examples, such as the pristine original (a high-score anchor) and images with entirely different distortion types from \emph{different source images}. This diverse context better defines the entire regression space and drives LSD's superior performance on objective tasks (\cref{tab:main_results}). However, this behavior also reveals a key limitation: for subjective preference tasks (e.g., AVA), this learned diversity introduces unnecessary variance, explaining why kNN's strict similarity approach remains superior.
\begin{figure}[t!]
    \centering
    \begin{subfigure}[b]{0.48\linewidth}
        \centering
        \includegraphics[width=\linewidth]{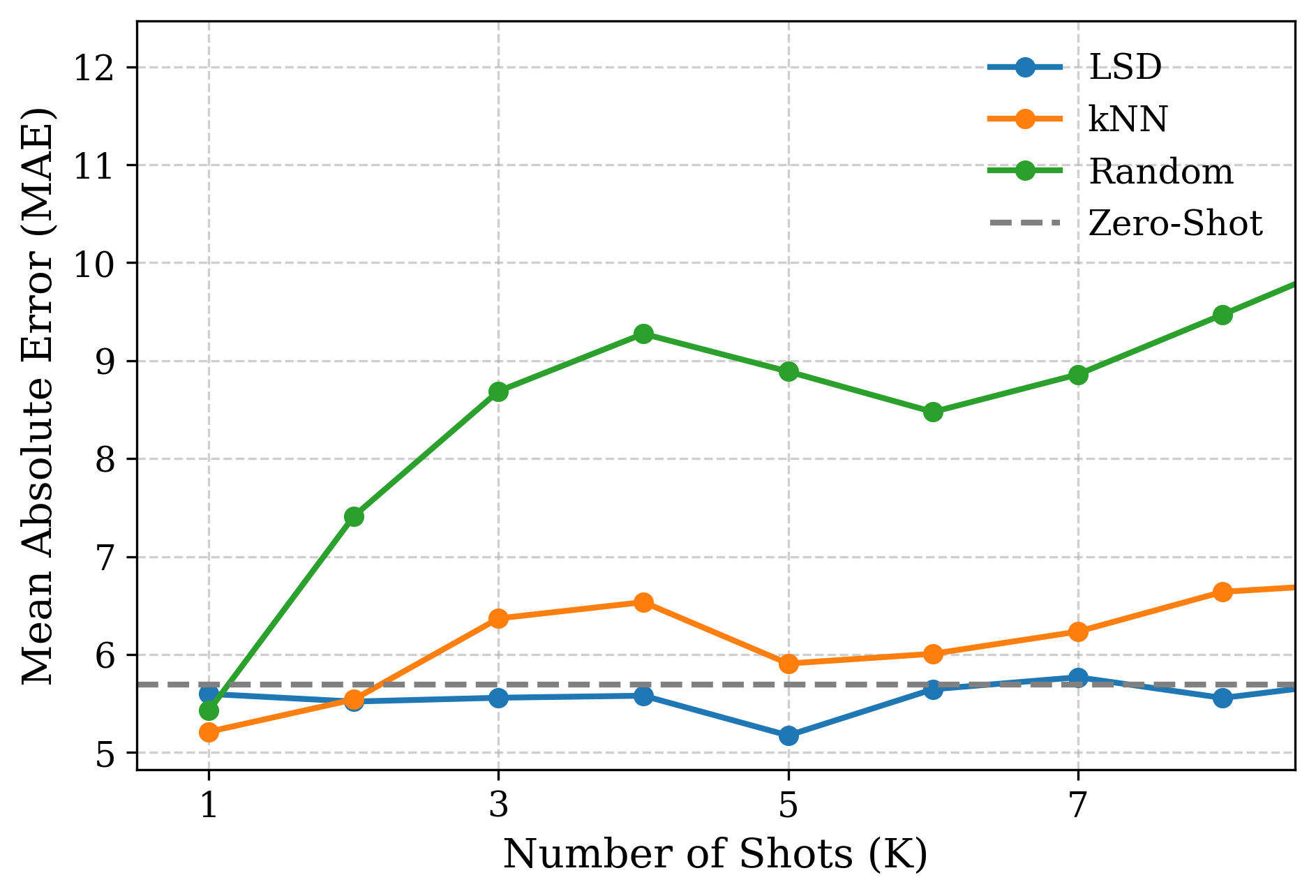}
        \caption{UTKFace on Qwen 2.5 7B}
        \label{fig:cross_vlm_qwen}
    \end{subfigure}
    \hfill 
    \begin{subfigure}[b]{0.48\linewidth}
        \centering
        \includegraphics[width=\linewidth]{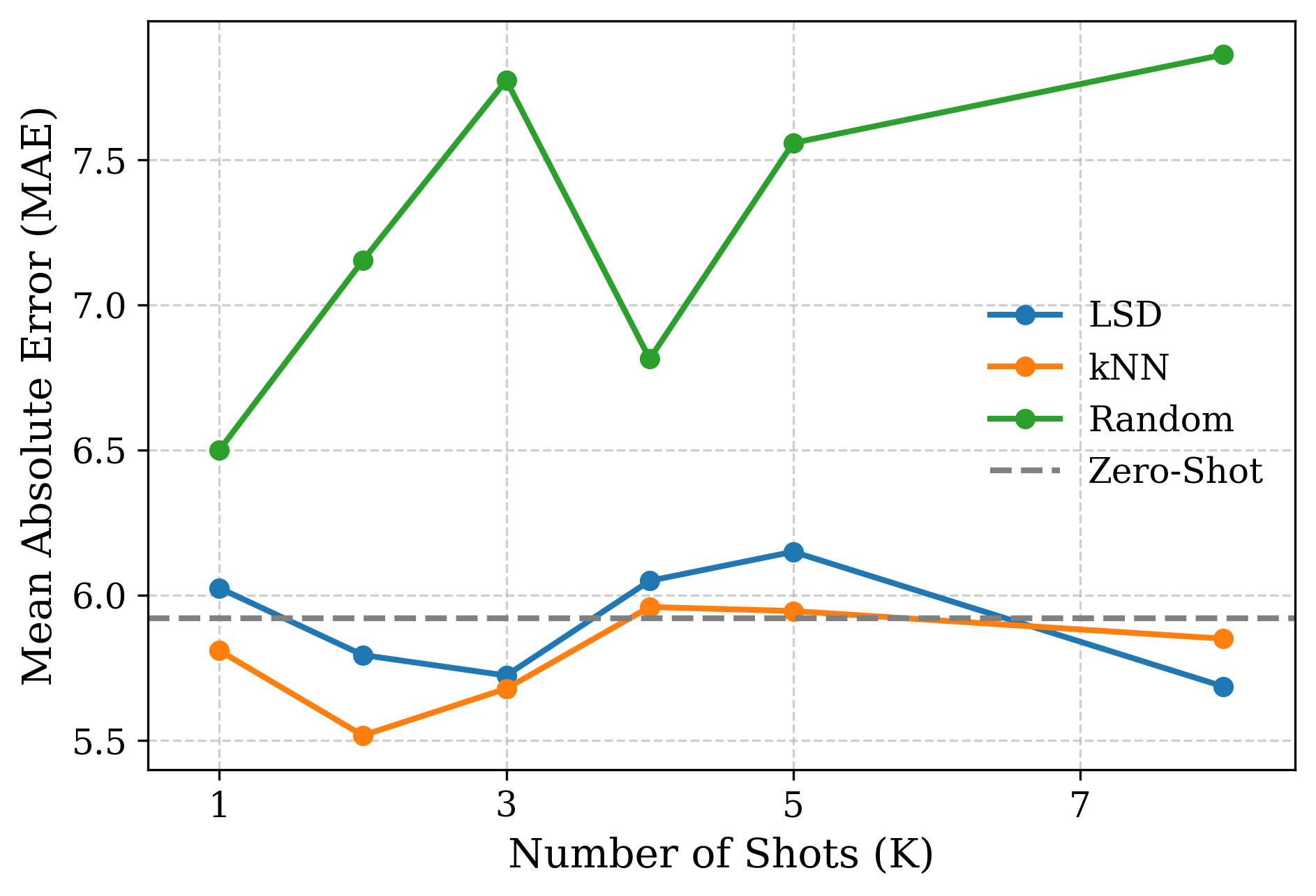}
        \caption{UTKFace on Phi-3.5-vision}
        \label{fig:cross_vlm_phi}
    \end{subfigure}
    
    \caption{
        \textbf{Cross-MLLM Generalization (MAE $\downarrow$) on UTKFace vs. Number of Shots ($K$).}
        We use the single LSD policy (trained on Gemma 3 4B-it) to select demos for two unseen MLLMs. The plots show our policy (blue line) versus the kNN (orange line) and Random (green line) baselines. 
        \textbf{(a)} On Qwen 2.5 7B, our policy consistently outperforms kNN. 
        \textbf{(b)} On Phi-3.5-vision, our policy performs on par with kNN. Both LSD and kNN significantly outperform the Random baseline.
    }
    \label{fig:cross_vlm}
\end{figure}

\subsubsection{Cross-MLLM Generalization}
This experiment tests whether the learned policy is MLLM-agnostic. We take the single LSD agent trained using rewards from Gemma 3 4B-it and use this \emph{frozen policy} to select demonstrations for \emph{Qwen 2.5 7B} and \emph{Phi-3.5-vision}. We evaluate performance on the objective UTKFace task, with results shown in \cref{fig:cross_vlm}.

The results show that our learned policy successfully transfers and remains highly effective, significantly outperforming the Random baseline on both models. The comparison to the strong kNN baseline is more nuanced. As shown in \cref{fig:cross_vlm}(a), our policy (blue line) maintains its performance advantage and consistently outperforms kNN (orange line) on Qwen 2.5 7B. On Phi-3.5-vision, shown in \cref{fig:cross_vlm}(b), our policy's performance is on par with kNN. This strongly suggests that our agent has learned a ``fundamental'' and generalizable policy that is not overfit to the original Gemma reward model, as it performs comparably or better than the strong kNN baseline on entirely unseen MLLMs.

\begin{table}[h!]
    \centering
    \caption{
        \textbf{Analysis of Selection Order (MAE $\downarrow$) at $K=8$.}
        We compare the performance of the agent's learned demonstration sequence against the exact same set of demonstrations in a random order.
    }
    \label{tab:order_analysis}
    \resizebox{\columnwidth}{!}{%
    \begin{tabular}{l|c|c}
        \toprule
        \textbf{Dataset} & \textbf{LSD (Learned Order)} & \textbf{LSD (Shuffled Set)} \\
        \midrule
        UTKFace      & 7.05 & \textbf{6.51} \\
        AVA          & \textbf{0.98} & 1.04 \\
        SCUT-FBP5500 & 0.67 & \textbf{0.53} \\
        KonIQ-10k    & \textbf{0.51} & 0.59 \\
        KADID-10k    & 0.82 & 0.82 \\
        \bottomrule
    \end{tabular}%
    }
\end{table}

\subsubsection{Analysis of Selection Order}
Our agent selects demonstrations $d_1, \dots, d_K$ sequentially. We sought to determine if this learned \emph{order} is a critical part of its policy, or if the agent is primarily learning to select a good \emph{set} of demonstrations. To test this, we conduct a permutation test. For each query, we first use our trained LSD agent to select its optimal $K=8$ demonstrations and record the MAE. Then, we randomly shuffle the order of those same 8 demonstrations and re-run inference.

As shown in \cref{tab:order_analysis}, the `Shuffled Set' performance is nearly identical to, and not consistently worse than, the agent's `Learned Order'. This strongly suggests that the primary skill our agent has learned is the selection of an optimal \emph{set} of demonstrations. The MLLM, in this case, appears robust to the permutation of those demonstrations, as long as the high-quality set is provided in the context.

\begin{table}[h!]
    \centering
    \caption{
        \textbf{Ablation Study on Decoder Input Strategy (MAE $\downarrow$).}
        We compare our query-centric model against a standard decoder-only model (Concat Input) on UTKFace for $K \in \{4, 8, 16\}$, both using $L=2$ layers. We also note the qualitative policy behavior.
    }
    \label{tab:ablation}
    \resizebox{\columnwidth}{!}{%
    \begin{tabular}{l|ccc|l}
        \toprule
        & \multicolumn{3}{c|}{\textbf{MAE $\downarrow$}} & \\
        \cmidrule(lr){2-4}
        \textbf{Decoder Input Strategy} & K=4 & K=8 & K=16 & \textbf{Policy Behavior} \\
        \midrule
        Query-Centric & \textbf{6.27} & 7.05 & \textbf{6.64} & Query-specific demos \\
        Concat Input & 7.01 & \textbf{6.42} & 7.74 & Non-query-specific \\
        \bottomrule
    \end{tabular}%
    }
\end{table}

\subsubsection{Ablation Study: State Encoder Architecture}
Our ablation study (\cref{tab:ablation}) compared our \emph{Query-Centric} model to a \emph{Concat Input} baseline, which concatenates all embeddings ($[\mathbf{e}_q; \mathbf{E}_{t-1}]$) as a single `tgt' sequence. The baseline exhibited a critical behavioral failure: \emph{policy collapse}, learning to select the same non-query-specific demonstrations for all queries. This fundamental failure confirms its inferiority, despite its inconsistent MAE scores. Our \emph{Query-Centric} model successfully learned a query-specific policy with strong and stable performance, proving our architectural choice is essential to learn an effective, non-degenerate policy.
\section{Conclusion}
\label{sec:conclusion}

We introduced LSD, a novel framework that reframes in-context demonstration selection as a sequential decision-making problem. Powered by a query-centric Transformer Decoder, our Dueling DQN agent learns a selection policy by optimizing for downstream MLLM performance, scaling to massive $O(N)$ action spaces via efficient FAISS-based retrieval. Crucially, our comprehensive evaluation reveals a fundamental task-dependent dichotomy in visual ICL: while simple kNN retrieval remains highly effective for subjective preference tasks, our learned policy is strictly necessary to achieve superior performance on objective visual regression tasks. By actively balancing visual relevance with necessary diversity, LSD develops an emergent awareness of the label structure to better define regression boundaries. This non-degenerate, generalizable policy demonstrates a clear path forward, illuminating exactly when learning—rather than simply retrieving—is essential for optimal in-context demonstrations.
\clearpage

\section*{Acknowledgements}
This work was supported by the National Institutes of Health (NIH R35GM128837).

{
    \small
    \bibliographystyle{ieeenat_fullname}
    \bibliography{main}

@String(CVPR= {IEEE Conf. Comput. Vis. Pattern Recog.})

@String(ICPR = {Int. Conf. Pattern Recog.})

@String(ICASSP=	{ICASSP})

@String(AAAI = {AAAI})

@String(CVPR  = {CVPR})

@String(ICPR  = {ICPR})

@inproceedings{purohitsample,
  title={Sample Efficient Demonstration Selection for In-Context Learning},
  author={Purohit, Kiran and Venktesh, V and Bhattacharya, Sourangshu and Anand, Avishek},
  booktitle={Forty-second International Conference on Machine Learning}
}

@article{zhang2023makes,
  title={What makes good examples for visual in-context learning?},
  author={Zhang, Yuanhan and Zhou, Kaiyang and Liu, Ziwei},
  journal={Advances in Neural Information Processing Systems},
  volume={36},
  pages={17773--17794},
  year={2023}
}

@inproceedings{yang2023representative,
  title={Representative demonstration selection for in-context learning with two-stage determinantal point process},
  author={Yang, Zhao and Zhang, Yuanzhe and Sui, Dianbo and Liu, Cao and Zhao, Jun and Liu, Kang},
  booktitle={Proceedings of the 2023 Conference on Empirical Methods in Natural Language Processing},
  pages={5443--5456},
  year={2023}
}

@article{ferber2024context,
  title={In-context learning enables multimodal large language models to classify cancer pathology images},
  author={Ferber, Dyke and W{\"o}lflein, Georg and Wiest, Isabella C and Ligero, Marta and Sainath, Srividhya and Ghaffari Laleh, Narmin and El Nahhas, Omar SM and M{\"u}ller-Franzes, Gustav and J{\"a}ger, Dirk and Truhn, Daniel and others},
  journal={Nature Communications},
  volume={15},
  number={1},
  pages={10104},
  year={2024},
  publisher={Nature Publishing Group UK London}
}

@inproceedings{doveh2024towards,
  title={Towards Multimodal In-context Learning for Vision and Language Models},
  author={Doveh, Sivan and Perek, Shaked and Mirza, M Jehanzeb and Lin, Wei and Alfassy, Amit and Arbelle, Assaf and Ullman, Shimon and Karlinsky, Leonid},
  booktitle={European Conference on Computer Vision},
  pages={250--267},
  year={2024},
  organization={Springer}
}

@inproceedings{liadvancing,
  title={Advancing Multimodal In-Context Learning in Large Vision-Language Models with Task-aware Demonstrations},
  author={Li, Yanshu},
  booktitle={Workshop on Reasoning and Planning for Large Language Models}
}

@article{xiao2025role,
  title={The Role of Diversity in In-Context Learning for Large Language Models},
  author={Xiao, Wenyang and Zhao, Haoyu and Huang, Lingxiao},
  journal={arXiv preprint arXiv:2505.19426},
  year={2025}
}

@article{yao2024large,
  title={Large language models are contrastive reasoners},
  author={Yao, Liang},
  journal={arXiv preprint arXiv:2403.08211},
  year={2024}
}

@article{dong2022survey,
  title={A survey on in-context learning},
  author={Dong, Qingxiu and Li, Lei and Dai, Damai and Zheng, Ce and Ma, Jingyuan and Li, Rui and Xia, Heming and Xu, Jingjing and Wu, Zhiyong and Liu, Tianyu and others},
  journal={arXiv preprint arXiv:2301.00234},
  year={2022}
}

@article{wei2022chain,
  title={Chain-of-thought prompting elicits reasoning in large language models},
  author={Wei, Jason and Wang, Xuezhi and Schuurmans, Dale and Bosma, Maarten and Xia, Fei and Chi, Ed and Le, Quoc V and Zhou, Denny and others},
  journal={Advances in neural information processing systems},
  volume={35},
  pages={24824--24837},
  year={2022}
}

@inproceedings{wang2023large,
  title={Large language models are implicitly topic models: Explaining and finding good demonstrations for in-context learning},
  author={Wang, Xinyi and Zhu, Wanrong and Saxon, Michael and Steyvers, Mark and Wang, William Yang},
  booktitle={Workshop on efficient systems for foundation models@ icml2023},
  year={2023}
}

@article{liu2024let,
  title={Let's Learn Step by Step: Enhancing In-Context Learning Ability with Curriculum Learning},
  author={Liu, Yinpeng and Liu, Jiawei and Shi, Xiang and Cheng, Qikai and Huang, Yong and Lu, Wei},
  journal={arXiv preprint arXiv:2402.10738},
  year={2024}
}

@article{liu2021makes,
  title={What Makes Good In-Context Examples for GPT-$3 $?},
  author={Liu, Jiachang and Shen, Dinghan and Zhang, Yizhe and Dolan, Bill and Carin, Lawrence and Chen, Weizhu},
  journal={arXiv preprint arXiv:2101.06804},
  year={2021}
}

@article{tanwar2023multilingual,
  title={Multilingual LLMs are better cross-lingual in-context learners with alignment},
  author={Tanwar, Eshaan and Dutta, Subhabrata and Borthakur, Manish and Chakraborty, Tanmoy},
  journal={arXiv preprint arXiv:2305.05940},
  year={2023}
}

@article{qin2023context,
  title={In-context learning with iterative demonstration selection},
  author={Qin, Chengwei and Zhang, Aston and Chen, Chen and Dagar, Anirudh and Ye, Wenming},
  journal={arXiv preprint arXiv:2310.09881},
  year={2023}
}

@article{rubin2021learning,
  title={Learning to retrieve prompts for in-context learning},
  author={Rubin, Ohad and Herzig, Jonathan and Berant, Jonathan},
  journal={arXiv preprint arXiv:2112.08633},
  year={2021}
}

@inproceedings{ye2023compositional,
  title={Compositional exemplars for in-context learning},
  author={Ye, Jiacheng and Wu, Zhiyong and Feng, Jiangtao and Yu, Tao and Kong, Lingpeng},
  booktitle={International Conference on Machine Learning},
  pages={39818--39833},
  year={2023},
  organization={PMLR}
}

@article{zhang2022active,
  title={Active example selection for in-context learning},
  author={Zhang, Yiming and Feng, Shi and Tan, Chenhao},
  journal={arXiv preprint arXiv:2211.04486},
  year={2022}
}

@article{kim2022self,
  title={Self-generated in-context learning: Leveraging auto-regressive language models as a demonstration generator},
  author={Kim, Hyuhng Joon and Cho, Hyunsoo and Kim, Junyeob and Kim, Taeuk and Yoo, Kang Min and Lee, Sang-goo},
  journal={arXiv preprint arXiv:2206.08082},
  year={2022}
}

@article{yang2023auto,
  title={Auto-icl: In-context learning without human supervision},
  author={Yang, Jinghan and Ma, Shuming and Wei, Furu},
  journal={arXiv preprint arXiv:2311.09263},
  year={2023}
}

@article{hao2022structured,
  title={Structured prompting: Scaling in-context learning to 1,000 examples},
  author={Hao, Yaru and Sun, Yutao and Dong, Li and Han, Zhixiong and Gu, Yuxian and Wei, Furu},
  journal={arXiv preprint arXiv:2212.06713},
  year={2022}
}

@article{an2023context,
  title={How do in-context examples affect compositional generalization?},
  author={An, Shengnan and Lin, Zeqi and Fu, Qiang and Chen, Bei and Zheng, Nanning and Lou, Jian-Guang and Zhang, Dongmei},
  journal={arXiv preprint arXiv:2305.04835},
  year={2023}
}

@article{wang2021want,
  title={Want to reduce labeling cost? GPT-3 can help},
  author={Wang, Shuohang and Liu, Yang and Xu, Yichong and Zhu, Chenguang and Zeng, Michael},
  journal={arXiv preprint arXiv:2108.13487},
  year={2021}
}

@article{khorashadizadeh2023exploring,
  title={Exploring in-context learning capabilities of foundation models for generating knowledge graphs from text},
  author={Khorashadizadeh, Hanieh and Mihindukulasooriya, Nandana and Tiwari, Sanju and Groppe, Jinghua and Groppe, Sven},
  journal={arXiv preprint arXiv:2305.08804},
  year={2023}
}

@article{ding2022gpt,
  title={Is gpt-3 a good data annotator?},
  author={Ding, Bosheng and Qin, Chengwei and Liu, Linlin and Chia, Yew Ken and Joty, Shafiq and Li, Boyang and Bing, Lidong},
  journal={arXiv preprint arXiv:2212.10450},
  year={2022}
}

@article{ram2023context,
  title={In-context retrieval-augmented language models},
  author={Ram, Ori and Levine, Yoav and Dalmedigos, Itay and Muhlgay, Dor and Shashua, Amnon and Leyton-Brown, Kevin and Shoham, Yoav},
  journal={Transactions of the Association for Computational Linguistics},
  volume={11},
  pages={1316--1331},
  year={2023},
  publisher={MIT Press One Broadway, 12th Floor, Cambridge, Massachusetts 02142, USA~…}
}

@inproceedings{panda2023differentially,
  title={Differentially private in-context learning},
  author={Panda, Ashwinee and Wu, Tong and Wang, Jiachen and Mittal, Prateek},
  booktitle={The 61st Annual Meeting Of The Association For Computational Linguistics},
  year={2023}
}

@article{meade2023using,
  title={Using in-context learning to improve dialogue safety},
  author={Meade, Nicholas and Gella, Spandana and Hazarika, Devamanyu and Gupta, Prakhar and Jin, Di and Reddy, Siva and Liu, Yang and Hakkani-T{\"u}r, Dilek},
  journal={arXiv preprint arXiv:2302.00871},
  year={2023}
}

@article{de2021editing,
  title={Editing factual knowledge in language models},
  author={De Cao, Nicola and Aziz, Wilker and Titov, Ivan},
  journal={arXiv preprint arXiv:2104.08164},
  year={2021}
}

@article{zhang2023sentiment,
  title={Sentiment analysis in the era of large language models: A reality check},
  author={Zhang, Wenxuan and Deng, Yue and Liu, Bing and Pan, Sinno Jialin and Bing, Lidong},
  journal={arXiv preprint arXiv:2305.15005},
  year={2023}
}

@article{xu2024improving,
  title={Improving in-context learning with prediction feedback for sentiment analysis},
  author={Xu, Hongling and Wang, Qianlong and Zhang, Yice and Yang, Min and Zeng, Xi and Qin, Bing and Xu, Ruifeng},
  journal={arXiv preprint arXiv:2406.02911},
  year={2024}
}

@inproceedings{wang2024context,
  title={In-context example retrieval from multi-perspectives for few-shot aspect-based sentiment analysis},
  author={Wang, Qianlong and Xu, Hongling and Ding, Keyang and Liang, Bin and Xu, Ruifeng},
  booktitle={Proceedings of the 2024 Joint International Conference on Computational Linguistics, Language Resources and Evaluation (LREC-COLING 2024)},
  pages={8975--8985},
  year={2024}
}

@article{yang2024empirical,
  title={An empirical study of Multimodal Entity-Based Sentiment Analysis with ChatGPT: Improving in-context learning via entity-aware contrastive learning},
  author={Yang, Li and Wang, Zengzhi and Li, Ziyan and Na, Jin-Cheon and Yu, Jianfei},
  journal={Information Processing \& Management},
  volume={61},
  number={4},
  pages={103724},
  year={2024},
  publisher={Elsevier}
}

@article{zhou2024visual,
  title={Visual in-context learning for large vision-language models},
  author={Zhou, Yucheng and Li, Xiang and Wang, Qianning and Shen, Jianbing},
  journal={arXiv preprint arXiv:2402.11574},
  year={2024}
}

@inproceedings{li2019deep,
  title={Deep instance-level hard negative mining model for histopathology images},
  author={Li, Meng and Wu, Lin and Wiliem, Arnold and Zhao, Kun and Zhang, Teng and Lovell, Brian},
  booktitle={International conference on medical image computing and computer-assisted intervention},
  pages={514--522},
  year={2019},
  organization={Springer}
}

@inproceedings{li2024context,
  title={In-context compositional generalization for large vision-language models},
  author={Li, Chuanhao and Jing, Chenchen and Li, Zhen and Zhai, Mingliang and Wu, Yuwei and Jia, Yunde},
  booktitle={Proceedings of the 2024 Conference on Empirical Methods in Natural Language Processing},
  pages={17954--17966},
  year={2024}
}

@inproceedings{zhu2025exploring,
  title={Exploring Task-Level Optimal Prompts for Visual In-Context Learning},
  author={Zhu, Yan and Ma, Huan and Zhang, Changqing},
  booktitle={Proceedings of the AAAI Conference on Artificial Intelligence},
  volume={39},
  number={10},
  pages={11031--11039},
  year={2025}
}

@article{guo2024makes,
  title={What Makes Good Few-shot Examples for Vision-Language Models?},
  author={Guo, Zhaojun and Lu, Jinghui and Liu, Xuejing and Zhao, Rui and Qian, ZhenXing and Tan, Fei},
  journal={arXiv preprint arXiv:2405.13532},
  year={2024}
}

@article{wu2025towards,
  title={Towards Reliable and Holistic Visual In-Context Learning Prompt Selection},
  author={Wu, Wenxiao and Xue, Jing-Hao and Xu, Chengming and Liu, Chen and Sun, Xinwei and Gao, Changxin and Sang, Nong and Fu, Yanwei},
  journal={arXiv preprint arXiv:2509.25989},
  year={2025}
}

@article{margatina2023active,
  title={Active learning principles for in-context learning with large language models},
  author={Margatina, Katerina and Schick, Timo and Aletras, Nikolaos and Dwivedi-Yu, Jane},
  journal={arXiv preprint arXiv:2305.14264},
  year={2023}
}

@article{zhang2025learning,
  title={Learning to Select In-Context Demonstration Preferred by Large Language Model},
  author={Zhang, Zheng and Lan, Shaocheng and Song, Lei and Bian, Jiang and Li, Yexin and Ren, Kan},
  journal={arXiv preprint arXiv:2505.19966},
  year={2025}
}

@article{wang2024demonstration,
  title={Demonstration selection for in-context learning via reinforcement learning},
  author={Wang, Xubin and Wu, Jianfei and Yuan, Yichen and Cai, Deyu and Li, Mingzhe and Jia, Weijia},
  journal={arXiv preprint arXiv:2412.03966},
  year={2024}
}

@inproceedings{zhifei2017cvpr,
  title={Age Progression/Regression by Conditional Adversarial Autoencoder},
  author={Zhang, Zhifei and Song, Yang and Qi, Hairong},
  booktitle={IEEE Conference on Computer Vision and Pattern Recognition (CVPR)},
  year={2017},
  organization={IEEE}
}

@article{hosu2020koniq,
  title={KonIQ-10k: An ecologically valid database for deep learning of blind image quality assessment},
  author={Hosu, Vlad and Lin, Hanhe and Sziranyi, Tamas and Saupe, Dietmar},
  journal={IEEE Transactions on Image Processing},
  volume={29},
  pages={4041--4056},
  year={2020},
  publisher={IEEE}
}

@inproceedings{lin2019kadid,
  title={KADID-10k: A large-scale artificially distorted IQA database},
  author={Lin, Hanhe and Hosu, Vlad and Saupe, Dietmar},
  booktitle={2019 Eleventh International Conference on Quality of Multimedia Experience (QoMEX)},
  pages={1--3},
  year={2019},
  organization={IEEE}
}

@inproceedings{liang2018scut,
  title={SCUT-FBP5500: A diverse benchmark dataset for multi-paradigm facial beauty prediction},
  author={Liang, Lingyu and Lin, Luojun and Jin, Lianwen and Xie, Duorui and Li, Mengru},
  booktitle={2018 24th International conference on pattern recognition (ICPR)},
  pages={1598--1603},
  year={2018},
  organization={IEEE}
}

@inproceedings{murray2012ava,
  title={AVA: A large-scale database for aesthetic visual analysis},
  author={Murray, Naila and Marchesotti, Luca and Perronnin, Florent},
  booktitle={2012 IEEE conference on computer vision and pattern recognition},
  pages={2408--2415},
  year={2012},
  organization={IEEE}
}

@inproceedings{paplham2024call,
  title={A call to reflect on evaluation practices for age estimation: Comparative analysis of the state-of-the-art and a unified benchmark},
  author={Paplh{\'a}m, Jakub and Franc, Vojt and others},
  booktitle={Proceedings of the IEEE/CVF Conference on Computer Vision and Pattern Recognition},
  pages={1196--1205},
  year={2024}
}

@article{xu2018transferring,
  title={Transferring rich deep features for facial beauty prediction},
  author={Xu, Lu and Xiang, Jinhai and Yuan, Xiaohui},
  journal={arXiv preprint arXiv:1803.07253},
  year={2018}
}

@inproceedings{tu2021regression,
  title={Regression or classification? new methods to evaluate no-reference picture and video quality models},
  author={Tu, Zhengzhong and Chen, Chia-Ju and Chen, Li-Heng and Wang, Yilin and Birkbeck, Neil and Adsumilli, Balu and Bovik, Alan C},
  booktitle={ICASSP 2021-2021 IEEE International Conference on Acoustics, Speech and Signal Processing (ICASSP)},
  pages={2085--2089},
  year={2021},
  organization={IEEE}
}

@inproceedings{wang2016dueling,
  title={Dueling network architectures for deep reinforcement learning},
  author={Wang, Ziyu and Schaul, Tom and Hessel, Matteo and Hasselt, Hado and Lanctot, Marc and Freitas, Nando},
  booktitle={International conference on machine learning},
  pages={1995--2003},
  year={2016},
  organization={PMLR}
}

@inproceedings{zhai2023sigmoid,
  title={Sigmoid loss for language image pre-training},
  author={Zhai, Xiaohua and Mustafa, Basil and Kolesnikov, Alexander and Beyer, Lucas},
  booktitle={Proceedings of the IEEE/CVF international conference on computer vision},
  pages={11975--11986},
  year={2023}
}

@article{douze2024faiss,
      title={The Faiss library},
      author={Matthijs Douze and Alexandr Guzhva and Chengqi Deng and Jeff Johnson and Gergely Szilvasy and Pierre-Emmanuel Mazaré and Maria Lomeli and Lucas Hosseini and Hervé Jégou},
      year={2024},
      eprint={2401.08281},
      archivePrefix={arXiv},
      primaryClass={cs.LG}
}

@article{team2025gemma,
  title={Gemma 3 technical report},
  author={Team, Gemma and Kamath, Aishwarya and Ferret, Johan and Pathak, Shreya and Vieillard, Nino and Merhej, Ramona and Perrin, Sarah and Matejovicova, Tatiana and Ram{\'e}, Alexandre and Rivi{\`e}re, Morgane and others},
  journal={arXiv preprint arXiv:2503.19786},
  year={2025}
}

@article{bai2025qwen2,
  title={Qwen2. 5-vl technical report},
  author={Bai, Shuai and Chen, Keqin and Liu, Xuejing and Wang, Jialin and Ge, Wenbin and Song, Sibo and Dang, Kai and Wang, Peng and Wang, Shijie and Tang, Jun and others},
  journal={arXiv preprint arXiv:2502.13923},
  year={2025}
}

@article{abdin2024phi,
  title={Phi-4 technical report},
  author={Abdin, Marah and Aneja, Jyoti and Behl, Harkirat and Bubeck, S{\'e}bastien and Eldan, Ronen and Gunasekar, Suriya and Harrison, Michael and Hewett, Russell J and Javaheripi, Mojan and Kauffmann, Piero and others},
  journal={arXiv preprint arXiv:2412.08905},
  year={2024}
}

@article{vaswani2017attention,
  title={Attention is all you need},
  author={Vaswani, Ashish and Shazeer, Noam and Parmar, Niki and Uszkoreit, Jakob and Jones, Llion and Gomez, Aidan N and Kaiser, {\L}ukasz and Polosukhin, Illia},
  journal={Advances in neural information processing systems},
  volume={30},
  year={2017}
}

@article{dulac2015deep,
  title={Deep reinforcement learning in large discrete action spaces},
  author={Dulac-Arnold, Gabriel and Evans, Richard and van Hasselt, Hado and Sunehag, Peter and Lillicrap, Timothy and Hunt, Jonathan and Mann, Timothy and Weber, Theophane and Degris, Thomas and Coppin, Ben},
  journal={arXiv preprint arXiv:1512.07679},
  year={2015}
}
}
\clearpage

\maketitlesupplementary
\appendix

\section{Prompt Construction and In-Context Learning}
\label{sec:prompt_construction}

We utilize a multimodal few-shot prompting strategy to query the Vision-Language Model (Gemma-3-4b-it). The prompt is constructed as an interleaved sequence of images and text, following the chat template structure required by the instruction-tuned model.

\subsection{Prompt Structure}
For a given query image $x_q$ and a selected set of $K$ demonstration examples $\{(x_1, y_1), (x_2, y_2), \dots, (x_K, y_K)\}$, where $x_i$ is an image and $y_i$ is its ground truth label (e.g., age, count, or score), the conversation history is constructed as follows:

\begin{equation}
\begin{split}
    \mathcal{P} = [ & I(x_1), T(y_1), I(x_2), T(y_2), \dots, \\
                    & I(x_K), T(y_K), I(x_q), Q_{task} ]
\end{split}
\end{equation}

\noindent where:
\begin{itemize}
    \item $I(x)$ represents the image token inputs processed by the vision encoder (SigLIP).
    \item $T(y)$ is the text string representing the label of the demonstration image (e.g., ``Age: 25'').
    \item $Q_{task}$ is the task-specific textual instruction that prompts the model to predict the label for the final image $x_q$.
\end{itemize}

We enforce strict output formatting by appending constraints to the system instruction and limiting generation to 20 tokens.

\subsection{Task-Specific Instructions}
The specific text prompts ($Q_{task}$) used for each dataset are detailed in Table~\ref{tab:prompts}.

\begin{table*}[t]
    \centering
    \caption{List of task-specific instructions used to prompt the VLM. The model is provided with $K$ interleaved image-label pairs prior to these instructions.}
    \label{tab:prompts}
    \renewcommand{\arraystretch}{1.2}
    \begin{tabular}{@{}l p{12cm}@{}}
        \toprule
        \textbf{Task / Dataset} & \textbf{Query Instruction ($Q_{task}$)} \\ 
        \midrule
        \textbf{Age Prediction} & ``What is the age of the person in the last image? Only output the estimated age as a number.'' \\
        \emph{(UTKFace)} & \\
        \midrule
        \textbf{Aesthetic Assessment} & ``What is the aesthetic score of the last image on a scale from 0 to 10? Only output the score as a floating number.'' \\
        \emph{(AVA)} & \\
        \midrule
        \textbf{Facial Beauty} & ``What is the facial beauty score of the last image on a scale from 0 to 5? Only output the score as a floating number.'' \\
        \emph{(FBP5500)} & \\
        \midrule
        \textbf{Image Quality} & ``What is the image quality score of the last image on a scale from 0 to 5? Only output the score as a floating number.'' \\
        \emph{(KADID-10k, KonIQ-10k)} & \\
        \bottomrule
    \end{tabular}
\end{table*}

\subsection{Label Formatting}
For the demonstration examples, the ground truth values are formatted as simple text strings to accompany the images:
\begin{itemize}
    \item \textbf{Age Prediction:} ``Age: $<y_i>$''
    \item \textbf{Scoring Tasks:} ``Score: $<y_i>$''
\end{itemize}
This consistency allows the VLM to recognize the mapping pattern effectively.

\section{Efficiency of Large-Scale Action Selection}
\label{sec:efficiency}

A critical challenge in applying Reinforcement Learning to retrieval tasks is the magnitude of the action space. In our setting, the agent must select from a dataset of $N \approx 50,000$ candidate images. Standard discrete RL algorithms (e.g., DQN, PPO) face prohibitive convergence and computational hurdles at this scale. We adopt a method inspired by the Wolpertinger architecture~\cite{dulac2015deep}, which maps states to continuous embedding coordinates rather than discrete indices. Below, we analyze the three primary advantages of this approach over discrete action spaces.

\subsection{Overcoming the Exploration Cliff}
In a standard discrete setting with $N$ actions, the policy $\pi$ must explore a multinomial distribution over $N$ independent logits.
\begin{itemize}
    \item \textbf{Discrete RL:} With $N=50,000$, the probability of selecting a specific optimal demonstration via random exploration (e.g., $\epsilon$-greedy) is $P(a^*) \approx 2 \times 10^{-5}$. This results in a vanishing gradient problem where the agent effectively never encounters a positive reward signal, leading to failure in convergence.
    \item \textbf{Proposed Method:} Our agent outputs a continuous vector $\hat{a} \in \mathbb{R}^D$. Exploration occurs in the semantic space. Even an imperfect vector output will retrieve neighbors in the embedding space that likely share task-relevant features with the optimal target, providing a denser reward signal and facilitating curriculum learning.
\end{itemize}

\subsection{Bridging the Semantic Gap}
Discrete RL treats actions as categorical indices without inherent relationships.
\begin{itemize}
    \item \textbf{Discrete RL:} To a standard DQN, index $i$ and index $j$ are orthogonal, even if the underlying images are semantically identical (e.g., two similar images of a ``Golden Retriever''). Learning that index $i$ yields high reward provides zero information about index $j$. The agent must independently explore and learn values for all $50,000$ indices.
    \item \textbf{Proposed Method:} By operating in the embedding space, we exploit the inductive bias of the pre-trained encoder (SigLIP). If the agent learns to navigate towards a specific region in $\mathbb{R}^D$ (e.g., ``dog-like images''), it simultaneously increases the selection probability for all semantically related candidates. This generalization capability drastically reduces the sample complexity required for training.
\end{itemize}
\subsection{Computational Complexity}
The computational cost of policy evaluation differs significantly between the methods due to the mechanism of action selection.

\begin{itemize}
    \item \textbf{Discrete RL:} Standard policy gradient methods require a Softmax normalization over the entire action space to compute probabilities:
    \begin{equation}
        P(a_i) = \frac{e^{z_i}}{\sum_{j=1}^{N} e^{z_j}}
    \end{equation}
    This incurs a computational complexity of $O(N)$ per step. As $N$ grows (e.g., $N=50,000$), calculating gradients for every output node becomes memory-intensive and prohibits efficient scaling.

    \item \textbf{Proposed Method (Two-Stage Selection):} Our approach decouples action generation from selection, reducing complexity from linear to logarithmic.
    \begin{enumerate}
        \item \textbf{Proto-Action Generation:} The policy outputs a continuous vector $\hat{a} \in \mathbb{R}^D$.
        \item \textbf{Candidate Retrieval:} We utilize Approximate Nearest Neighbor search (FAISS) to retrieve a small set of candidates $\mathcal{C}_k$ (where $k \ll N$, e.g., $k=200$) closest to $\hat{a}$. This search scales logarithmically $O(\log N)$.
        \item \textbf{Final Selection:} The policy evaluates Q-values only for the actions in $\mathcal{C}_k$. 
    \end{enumerate}
    The total complexity is $O(\log N + k)$, which allows the method to scale to millions of images while keeping the evaluation cost constant.
\end{itemize}

\begin{table}[h]
    \centering
    \caption{Comparison between Standard Discrete RL and the proposed Continuous Embedding (Wolpertinger) approach for large-scale selection.}
    \label{tab:efficiency_comparison}
    \resizebox{\linewidth}{!}{
    \begin{tabular}{@{}lll@{}}
        \toprule
        \textbf{Feature} & \textbf{Discrete RL (e.g., PPO)} & \textbf{Proposed Method} \\ 
        \midrule
        \textbf{Output Space} & $N$ logits (One per item) & Vector $\in \mathbb{R}^D$ \\
        \textbf{Selection} & Softmax over $N$ & ANN Retrieval ($k$) $\to$ Argmax \\
        \textbf{Semantics} & Orthogonal Actions & Semantic Neighbors \\
        \textbf{Complexity} & Linear $O(N)$ & Logarithmic $O(\log N + k)$ \\
        \textbf{Scalability} & Limited ($N < 10^4$) & Unlimited (via FAISS) \\
        \bottomrule
    \end{tabular}
    }
\end{table}

\section{Additional Implementation Details}
\subsection{Network Architecture \& Hyperparameters}
Our Dueling DQN agent utilizes a custom \textbf{Query-Centric Transformer Decoder} with specific architectural choices designed for stability and sample efficiency. The exact configuration is detailed below:

\begin{itemize}
    \item \textbf{Transformer Configuration:}
    \begin{itemize}
        \item \textbf{Layers ($L$):} 2
        \item \textbf{Attention Heads ($H$):} 4
        \item \textbf{Embedding Dimension ($D$):} 768 (matching the SigLIP vision encoder)
        \item \textbf{Feedforward Dimension:} 3072 ($4 \times D$)
        \item \textbf{Positional Encoding:} Learnable embeddings are added to the demonstration memory sequence to encode slot order.
        \item \textbf{Normalization:} LayerNorm is applied \emph{before} the attention/FFN blocks (\texttt{norm\_first=True}) for improved training stability.
        \item \textbf{Activation:} GELU
        \item \textbf{Dropout:} 0.1
    \end{itemize}
    
    \item \textbf{Dueling Heads:} 
    Unlike standard architectures that use heavy MLPs, we found that single linear projections were sufficient given the rich representations from the Transformer context.
    \begin{itemize}
        \item \textbf{Value Head:} A single linear layer mapping $D \to 1$.
        \item \textbf{Advantage Head:} A single linear layer mapping $D \to D$, followed by $L_2$ normalization.
    \end{itemize}
\end{itemize}

\subsection{Optimization Process}
We train the agent using the \textbf{Adam} optimizer with a learning rate of $5 \times 10^{-6}$. We employ gradient clipping with a max norm of $1.0$ to ensure stability throughout the training process. The target network is updated using a soft update parameter $\tau = 0.005$.

We utilize a Replay Buffer with a capacity of 50,000 transitions and sample mini-batches of size 32. To encourage exploration, we use an $\epsilon$-greedy schedule starting at $\epsilon=0.9$ and decaying exponentially to $\epsilon=0.05$ over the first 100,000 steps.

\subsection{Data Splits and Evaluation Protocol}
\label{sec:data_split}

To ensure rigorous evaluation and prevent data leakage, we perform a strict separation between the data used for demonstration retrieval and the data used for evaluation queries.

\subsubsection{Dataset Partitioning}
For each task (e.g., Age Estimation, Aesthetic Scoring), we first load the raw dataset and filter for valid images. To maintain a manageable memory footprint for the FAISS index during experimentation, we cap the maximum dataset size at $N_{max} = 25,000$ samples. If a dataset exceeds this limit, we perform a random downsampling.

We partition this data into two disjoint sets using an 80/20 random split:
\begin{itemize}
    \item \textbf{Demonstration Pool ($\mathcal{D}_{train}$):} Comprising 80\% of the data, this set serves as the candidate pool. All $K$-shot demonstrations retrieved by our agent (or baselines) are strictly drawn from this pool. The FAISS index is built solely on the SigLIP embeddings of this set.
    \item \textbf{Query Set ($\mathcal{D}_{test}$):} Comprising the remaining 20\%, this set is used exclusively to provide query images $x_q$. These images are never used as demonstrations.
\end{itemize}

\subsubsection{Evaluation Sampling}
While the Query Set ($\mathcal{D}_{test}$) may contain up to 5,000 images (20\% of 25,000), performing full inference on large Vision-Language Models (e.g., Gemma-3-4B-IT, InternVL2) for every sample is computationally prohibitive due to the latency of auto-regressive generation.

To balance statistical significance with computational efficiency, we randomly sample a fixed subset of $N_{eval} = 1,000$ queries from $\mathcal{D}_{test}$ for the final quantitative evaluation. This sample size is sufficient to capture performance trends and calculate metrics (MAE, Accuracy) with low variance while keeping the evaluation time feasible.

\subsection{FAISS Index Configuration and Retrieval Strategy}
To efficiently handle the action space of $N \approx 50,000$ images, we employ the Facebook AI Similarity Search (FAISS) library. We construct an Inverted File with Product Quantization (`IndexIVFPQ') index to balance memory usage with retrieval speed. The configuration matches the embedding dimension of the SigLIP encoder ($D=768$).

\begin{itemize}
    \item \textbf{Metric:} Inner Product. We use \texttt{METRIC\_INNER\_}\-\texttt{PRODUCT}. Since all embeddings are $L_2$ normalized, this is equivalent to Cosine Similarity.
    \item \textbf{Coarse Quantizer (`nlist'):} 100 Voronoi cells. We use a flat inner product quantizer (`IndexFlatIP`) for the coarse level.
    \item \textbf{Sub-Quantizers (`M'):} 8. The vectors are split into 8 sub-vectors.
    \item \textbf{Encoding Bits:} 8 bits per sub-vector.
    \item \textbf{Search Depth (`nprobe'):} 10. During inference and training, we visit the nearest 10 Voronoi cells.
    \item \textbf{Candidate Pool Size ($k$):} 200. During the training step, we retrieve the top $k=200$ candidates for the generated proto-action.  This pool size is critical for the Dueling DQN architecture, as it serves as the sample set to approximate the mean advantage value: 
    \begin{equation}
        \bar{A}(s, \cdot) \approx \frac{1}{k} \sum_{a_j \in \text{top-}k} A(s, a_j)
    \end{equation}
\end{itemize}

\subsection{Environment \& Reward Shaping}
To address the cold-start problem, the environment employs an \textbf{anchor initialization} strategy: the initial state $s_0$ always includes the query image and its nearest neighbor (retrieved via FAISS) as the first demonstration.

We utilize a \textbf{differential reward function} to encourage marginal improvement. The immediate reward $r_t$ is calculated as the change in the MLLM's performance score:
\begin{equation}
    r_t = \frac{1}{\lambda} (S_t - S_{t-1})
\end{equation}
\noindent where $\lambda=10.0$ is a global scaling constant for the replay buffer. 

The performance score $S_t$ is task-dependent, designed to normalize the error magnitude across different output ranges (e.g., $0-100$ for age vs. $0-5$ for quality). Let $\delta = |y_{pred} - y_{gt}|$ be the absolute error. $S_t$ is defined as:

\begin{equation}
    S_t = 
    \begin{cases} 
        -\delta & \text{Age Prediction} \\
        -10 \cdot \delta & \text{Aesthetic Scoring (0-10)} \\
        -20 \cdot \delta & \text{Quality/Beauty Scoring (0-5)}
    \end{cases}
\end{equation}

\noindent For crowd counting, we use a relative error formulation (damped by a floor of 10 heads) to handle the large variance in crowd sizes. For scoring tasks, we apply multipliers (10 or 20) to amplify the gradient signal for small floating-point errors.
If the agent selects an invalid action, the episode terminates with a fixed penalty of $-0.5$.

\section{Extended Demonstration Set Analysis}
\label{sec:supp_extended_set}

\begin{figure*}[p]
    \centering
    \small
    \textbf{COLUMNS: (Left) Demo-Query Similarity; (Right) Pairwise Similarity} \par \vspace{0.5em}
    \begin{tabular}{c c}
        \begin{subfigure}[b]{0.3\linewidth}
            \includegraphics[width=\linewidth]{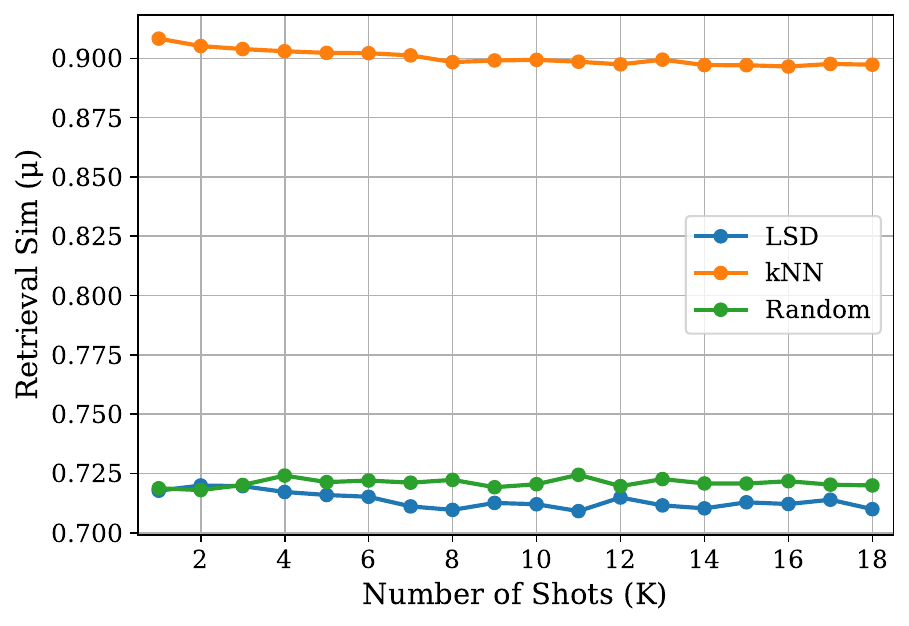}
            \caption*{UTKFace: Demo-Query Feature Similarity}
        \end{subfigure}
        & 
        \begin{subfigure}[b]{0.3\linewidth}
            \includegraphics[width=\linewidth]{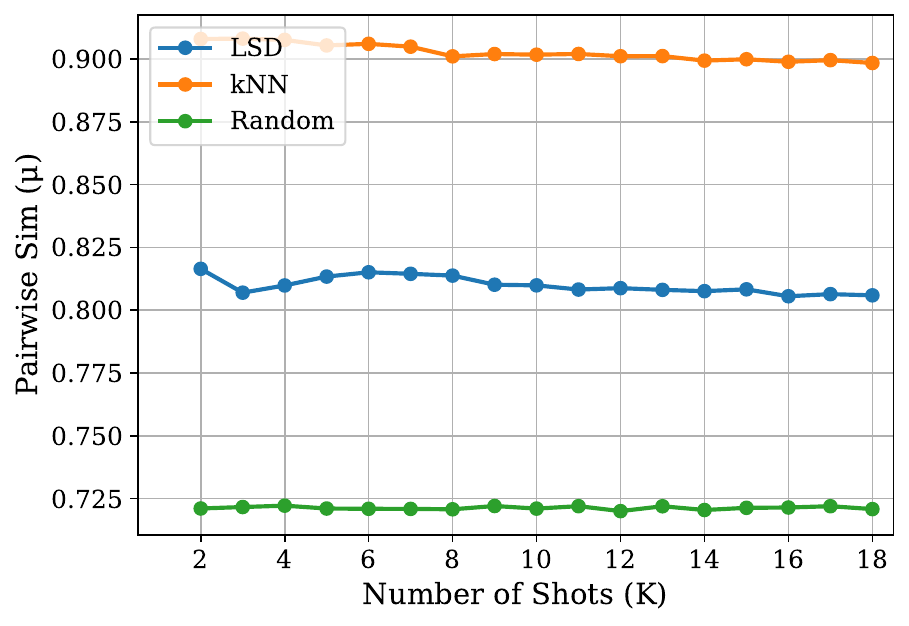}
            \caption*{UTKFace: Pairwise Feature Similarity}
        \end{subfigure}
        \\ 
        \begin{subfigure}[b]{0.3\linewidth}
            \includegraphics[width=\linewidth]{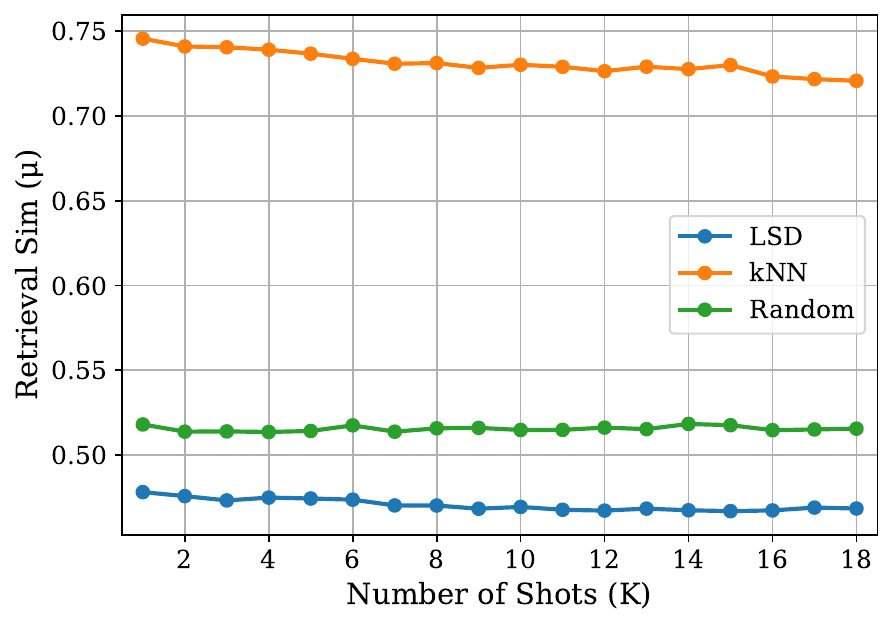}
            \caption*{AVA: Demo-Query Feature Similarity}
        \end{subfigure}
        &
        \begin{subfigure}[b]{0.3\linewidth}
            \includegraphics[width=\linewidth]{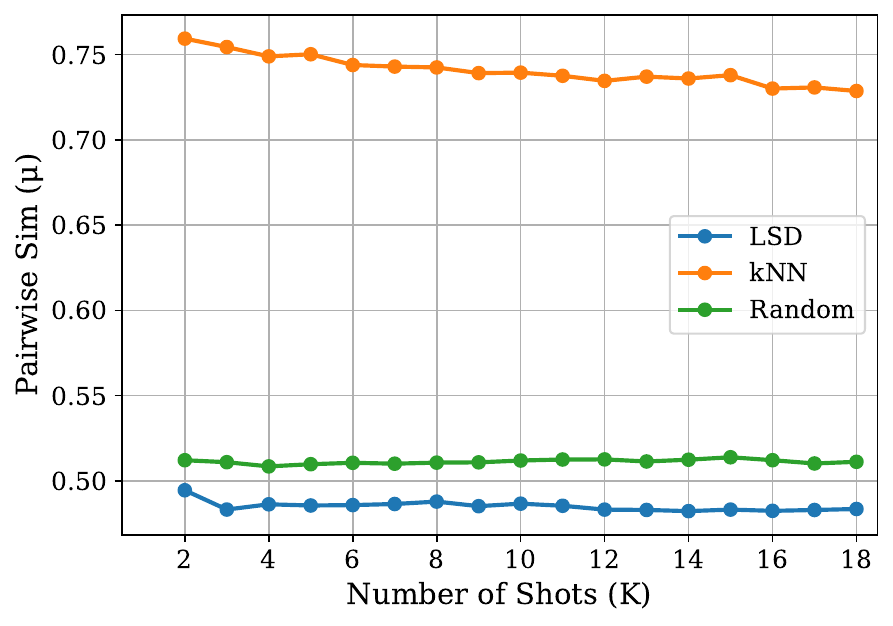}
            \caption*{AVA: Pairwise Feature Similarity}
        \end{subfigure}
        \\
        \begin{subfigure}[b]{0.3\linewidth}
            \includegraphics[width=\linewidth]{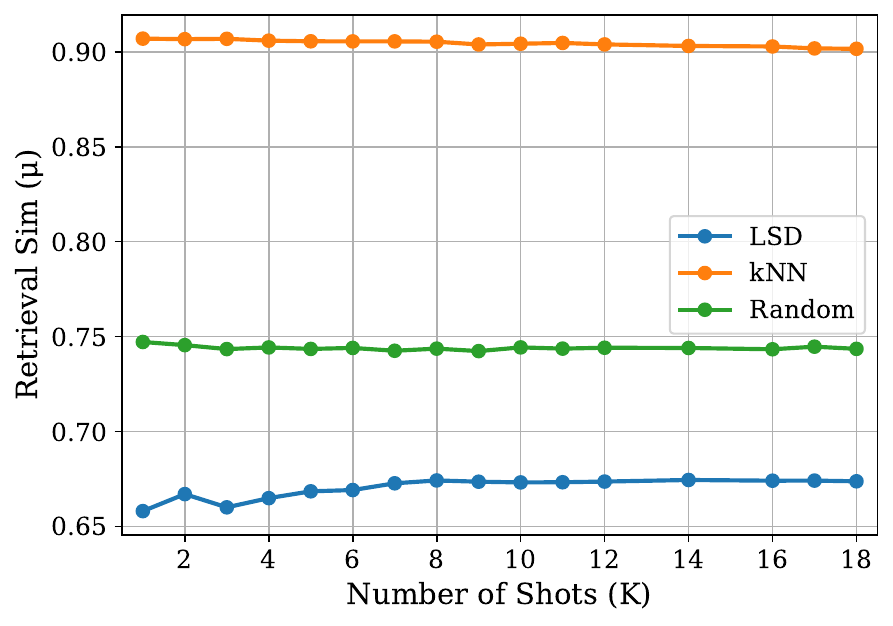}
            \caption*{SCUT-FBP5500: Demo-Query Feature Similarity}
        \end{subfigure}
        &
        \begin{subfigure}[b]{0.3\linewidth}
            \includegraphics[width=\linewidth]{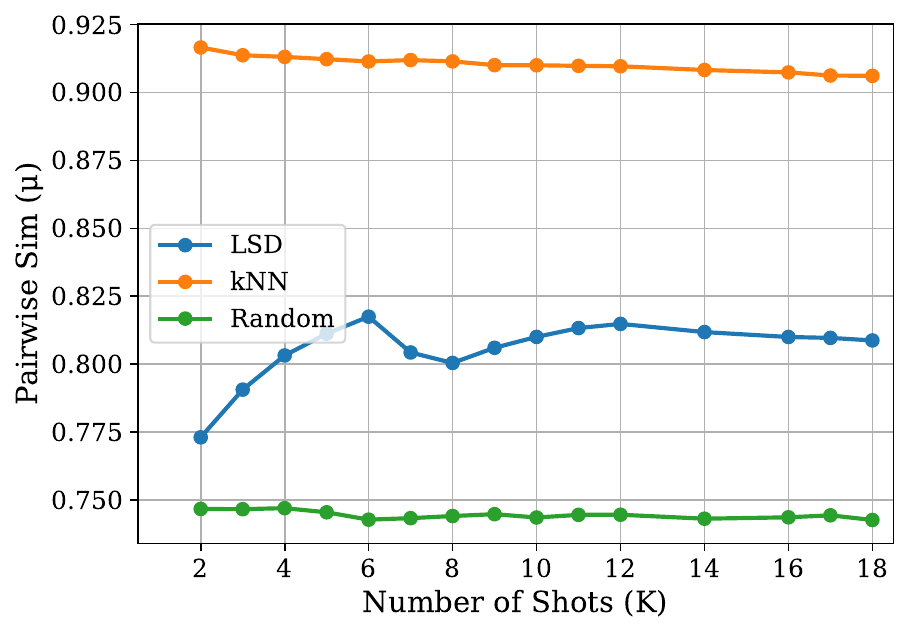}
            \caption*{SCUT-FBP5500: Pairwise Feature Similarity}
        \end{subfigure}
        \\
        \begin{subfigure}[b]{0.3\linewidth}
            \includegraphics[width=\linewidth]{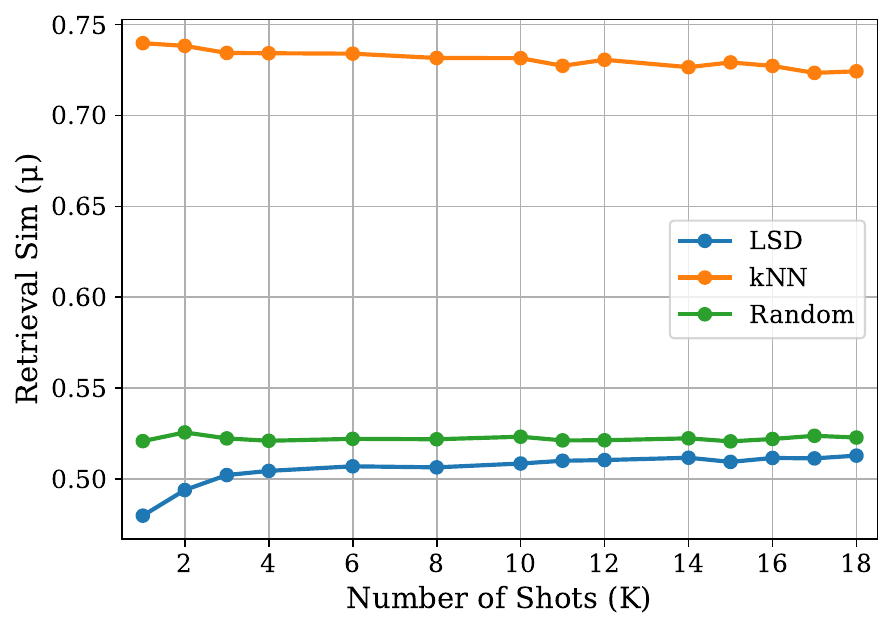}
            \caption*{KonIQ-10k: Demo-Query Feature Similarity}
        \end{subfigure}
        &
        \begin{subfigure}[b]{0.3\linewidth}
            \includegraphics[width=\linewidth]{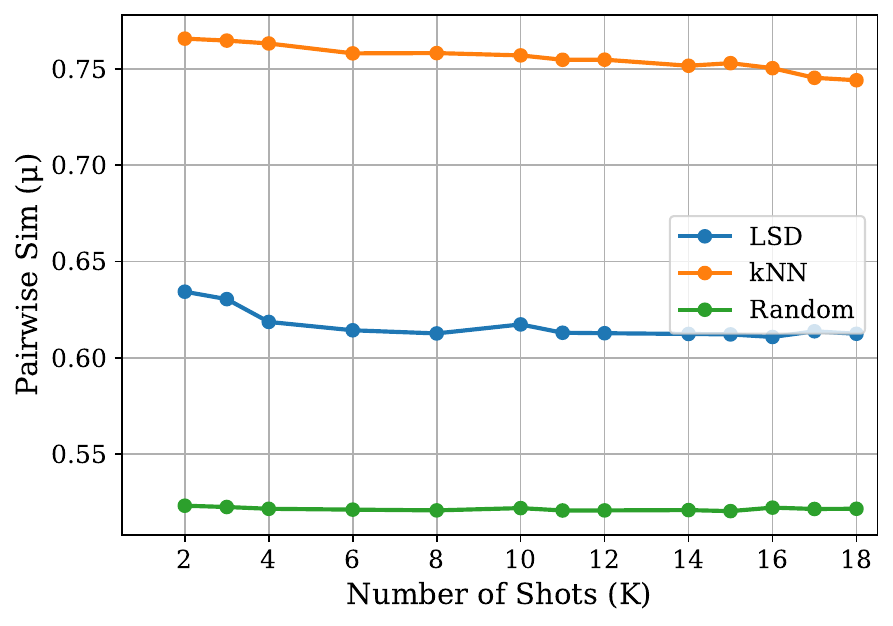}
            \caption*{KonIQ-10k: Pairwise Feature Similarity}
        \end{subfigure}
        \\
        \begin{subfigure}[b]{0.3\linewidth}
            \includegraphics[width=\linewidth]{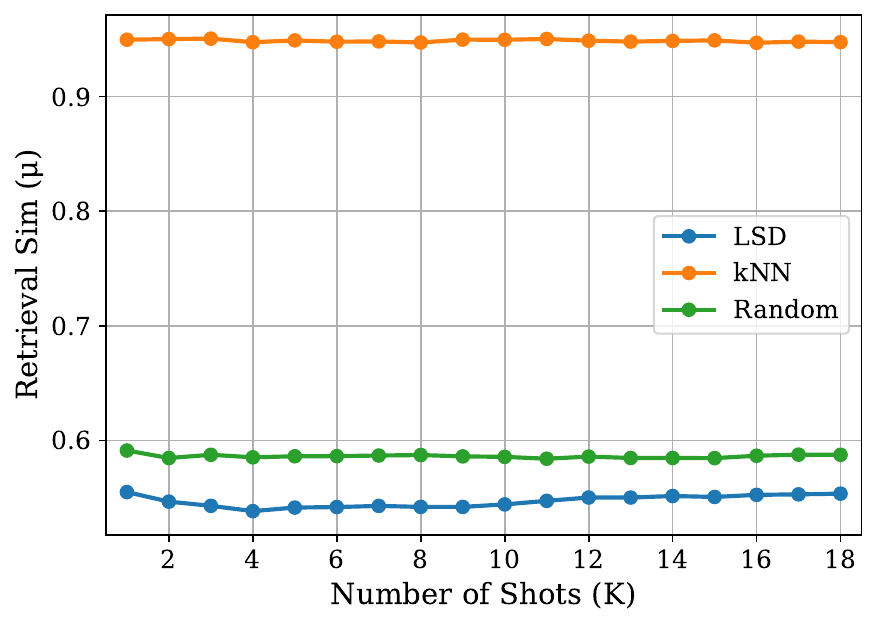}
            \caption*{KADID-10k: Demo-Query Feature Similarity}
        \end{subfigure}
        &
        \begin{subfigure}[b]{0.3\linewidth}
            \includegraphics[width=\linewidth]{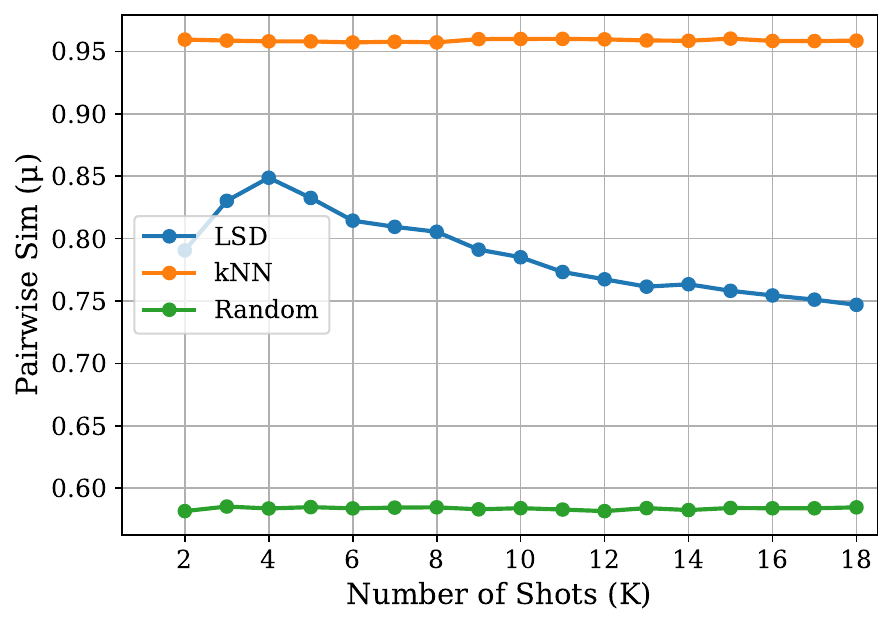}
            \caption*{KADID-10k: Pairwise Feature Similarity}
        \end{subfigure}
    \end{tabular}
    \caption{
        \textbf{Extended Feature-Space Analysis (Relevance and Similarity).}
        The plots in the right column demonstrate that on all five datasets, LSD (blue line) actively seeks low redundancy, maintaining the trend $\text{LSD} \ll \text{kNN}$ in pairwise similarity, which is the key behavioral difference.
    }
    \label{fig:supp_diversity_analysis}
\end{figure*}

\begin{figure*}[p] 
    \centering
    \small
    \textbf{COLUMNS: (Left) Label MAE vs. Query $\downarrow$; (Right) Pairwise Label MAE $\downarrow$} \par \vspace{0.5em}
    \begin{tabular}{c c}
        \begin{subfigure}[b]{0.3\linewidth}
            \centering
            \includegraphics[width=\linewidth]{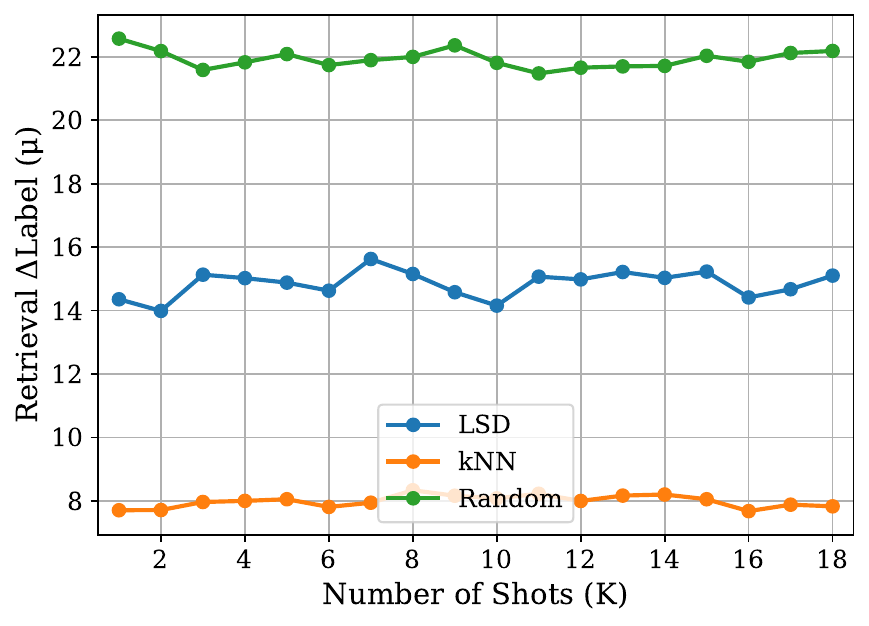}
            \caption{UTKFace: Label MAE vs. Query} 
        \end{subfigure}
        &
        \begin{subfigure}[b]{0.3\linewidth}
            \centering
            \includegraphics[width=\linewidth]{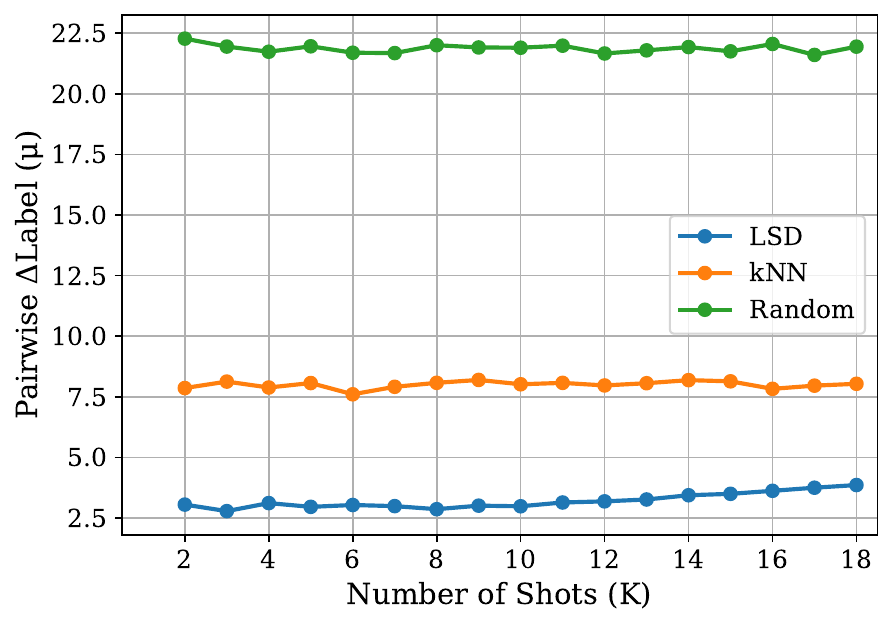}
            \caption{UTKFace: Pairwise Label MAE}
        \end{subfigure}
        \\
        \begin{subfigure}[b]{0.3\linewidth}
            \centering
            \includegraphics[width=\linewidth]{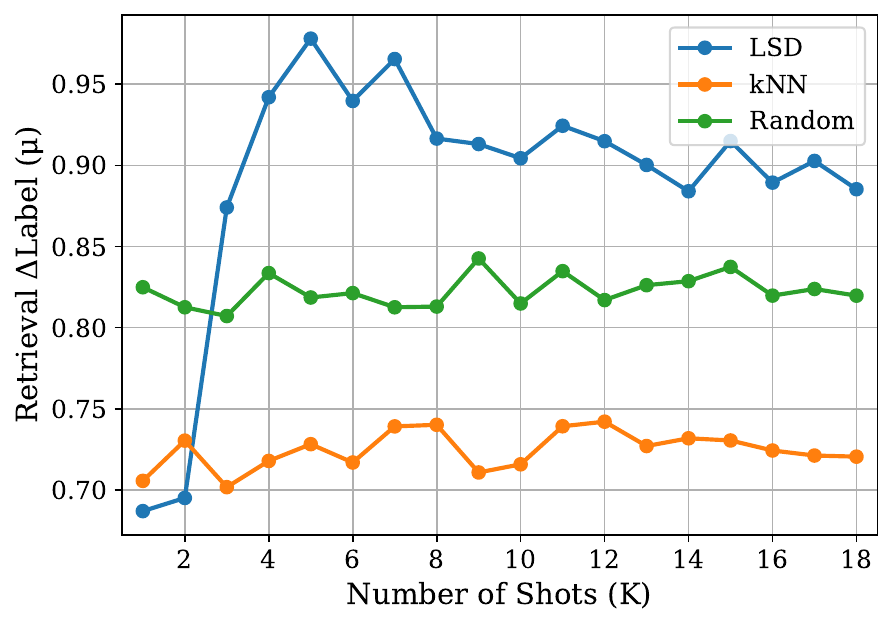}
            \caption{AVA: Label MAE vs. Query}
        \end{subfigure}
        &
        \begin{subfigure}[b]{0.3\linewidth}
            \centering
            \includegraphics[width=\linewidth]{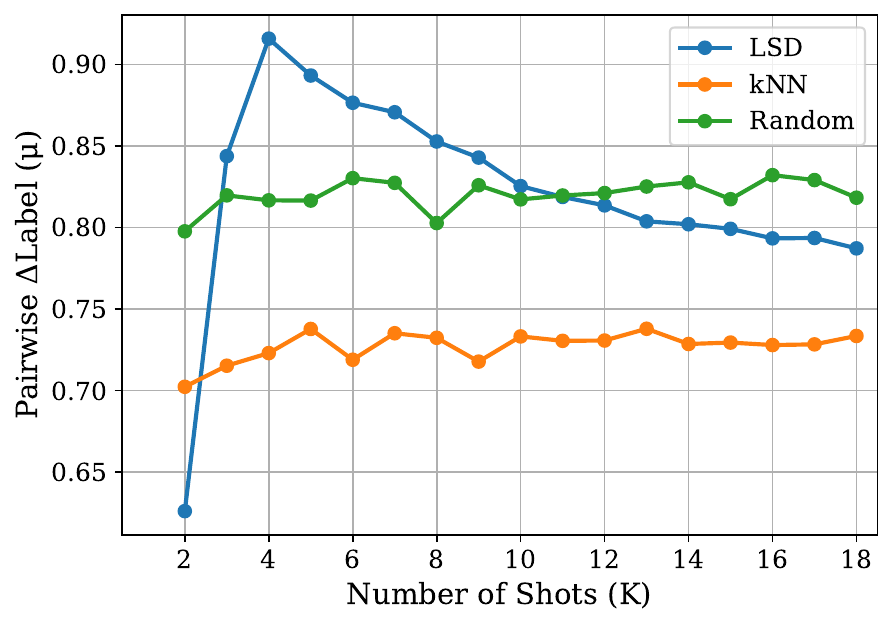}
            \caption{AVA: Pairwise Label MAE}
        \end{subfigure}
        \\
        \begin{subfigure}[b]{0.3\linewidth}
            \centering
            \includegraphics[width=\linewidth]{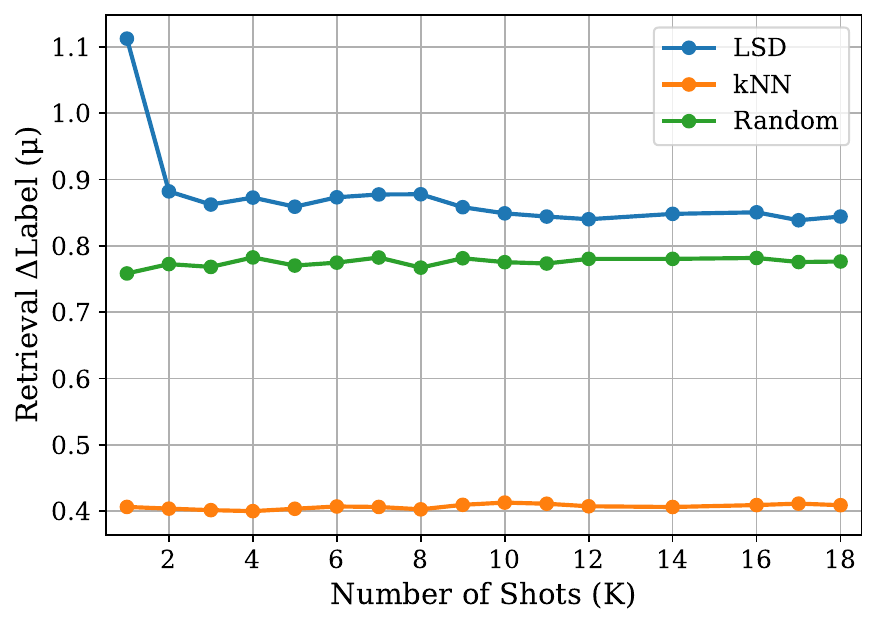}
            \caption{SCUT-FBP5500: Label MAE vs. Query}
        \end{subfigure}
        &
        \begin{subfigure}[b]{0.3\linewidth}
            \centering
            \includegraphics[width=\linewidth]{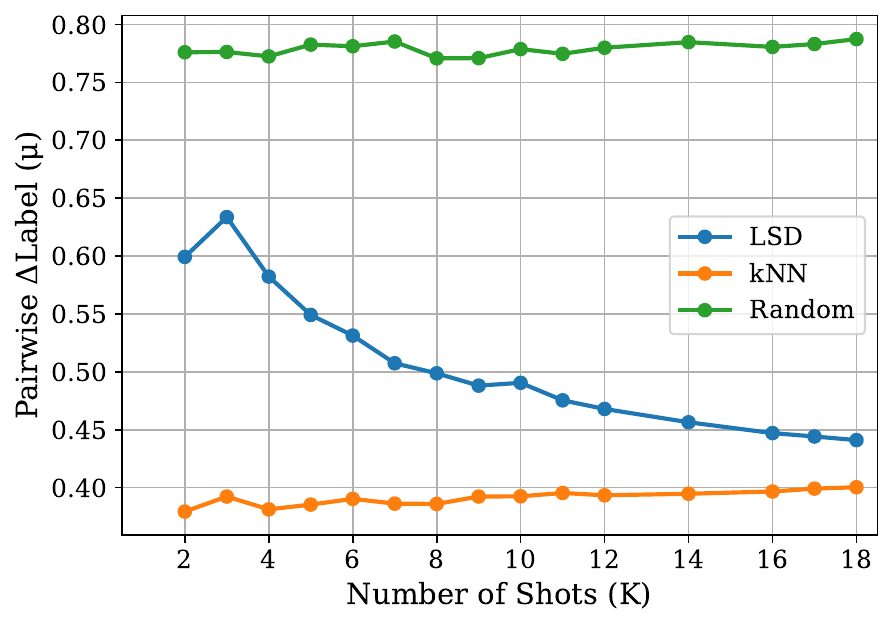}
            \caption{SCUT-FBP5500: Pairwise Label MAE}
        \end{subfigure}
        \\
        \begin{subfigure}[b]{0.3\linewidth}
            \centering
            \includegraphics[width=\linewidth]{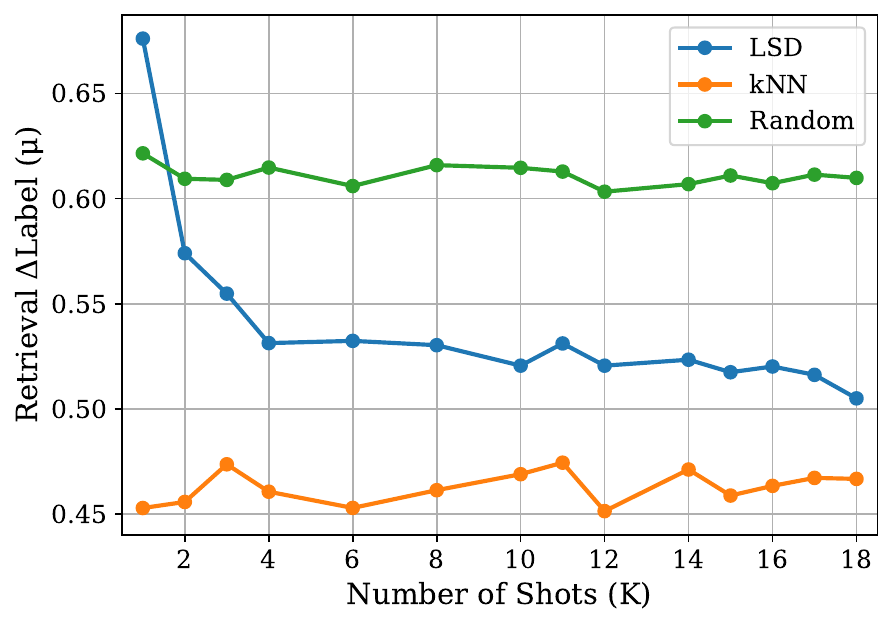}
            \caption{KonIQ-10k: Label MAE vs. Query}
        \end{subfigure}
        &
        \begin{subfigure}[b]{0.3\linewidth}
            \centering
            \includegraphics[width=\linewidth]{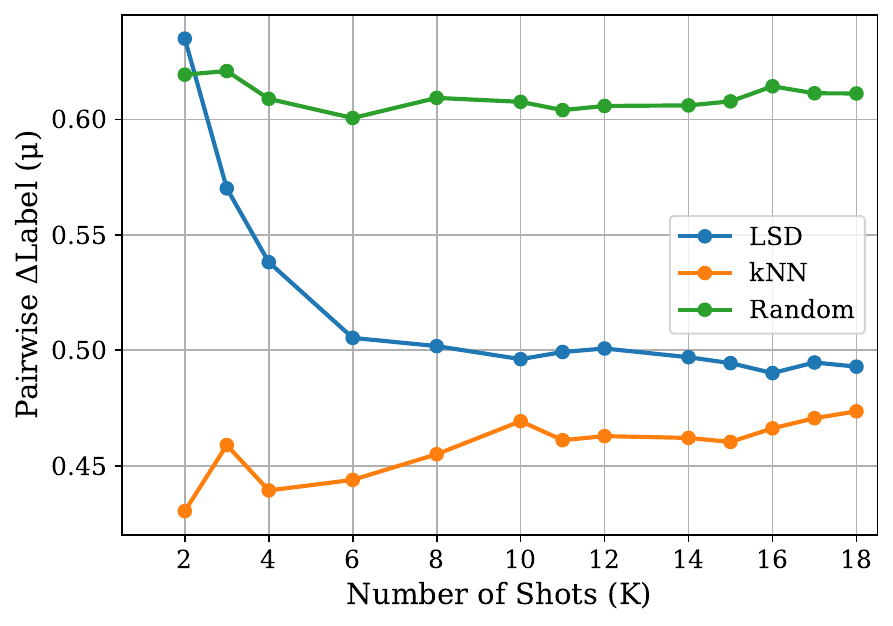}
            \caption{KonIQ-10k: Pairwise Label MAE}
        \end{subfigure}
        \\
        \begin{subfigure}[b]{0.3\linewidth}
            \centering
            \includegraphics[width=\linewidth]{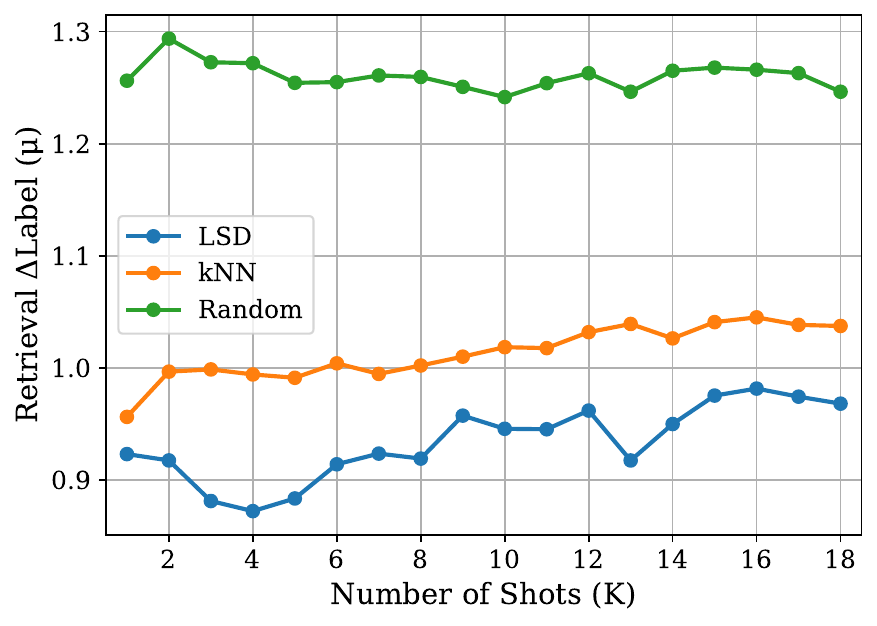}
            \caption{KADID-10k: Label MAE vs. Query}
        \end{subfigure}
        &
        \begin{subfigure}[b]{0.3\linewidth}
            \centering
            \includegraphics[width=\linewidth]{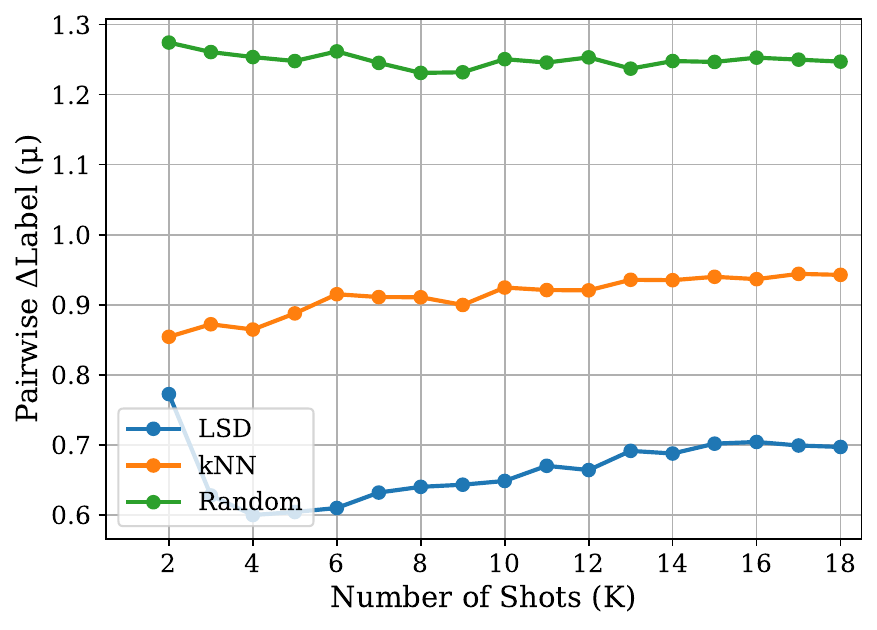}
            \caption{KADID-10k: Pairwise Label MAE}
        \end{subfigure}
    \end{tabular}
    \caption{
        \textbf{Extended Label-Space Analysis (Emergent Relevance and Consistency).}
        The results show a critical, task-dependent pattern in minimizing the label difference ($\Delta$Label) between the query and selected demos. For \textbf{Objective Tasks} (UTKFace, KonIQ-10k, KADID-10k), the \textbf{LSD policy (blue)} is the most effective implicit label retriever. Conversely, for \textbf{Subjective Tasks} (AVA, SCUT-FBP5500), the \textbf{kNN Baseline (orange)} consistently selects demos that minimize the label difference. This confirms that the optimal policy for minimizing label-space MAE aligns with the task's underlying human perception structure.
    }
    \label{fig:supp_relevance_analysis}
\end{figure*}

To ensure the universality of our learned policy, we extended the Demonstration Set Analysis presented in the main paper to cover all five benchmark datasets. This experiment was conducted in the Intra-Model setting (Train Gemma 3 4B-it / Eval Gemma 3 4B-it). The results, summarized visually in \cref{fig:supp_diversity_analysis} and \cref{fig:supp_relevance_analysis}, consistently confirm the emergence of a sophisticated, multi-objective policy across all tasks.

\section{Comprehensive Cross-Model Generalization}

In the main paper, we demonstrated the transfer capability of a single policy. Here, we present a comprehensive, all-to-all transfer analysis to rigorously test the universality of our approach.

We utilize three state-of-the-art MLLMs for training source policies: \textbf{Gemma 3 4B-it}, \textbf{Qwen 2.5 7B}, and \textbf{InternVL2-8B}. We evaluate these policies against all four target MLLMs, including \textbf{Phi-3.5-vision}. We define a transfer matrix experiment where we train a distinct LSD agent on each source model and evaluate it on every target model.

\subsection{The Transfer Matrix}
The aggregate results on the UTKFace dataset ($K=4$) are presented in \cref{tab:transfer_matrix}.

\begin{table*}[h!]
    \centering
    \caption{
        \textbf{Cross-Model Transfer Matrix (MAE $\downarrow$) on UTKFace at $K=4$.}
        Rows represent the \emph{Source Policy} (training model). Columns represent the \emph{Target MLLM} (evaluation model).
        \textbf{Diagonal entries (gray)} represent Intra-Model performance (Specialization), where LSD typically achieves its peak performance.
        \textbf{Off-diagonal entries} represent Inter-Model performance (Generalization).
        While LSD consistently outperforms Random selection (not shown), it performs comparably to the kNN baseline in cross-model settings, suggesting that model-specific nuances play a significant role in optimal retrieval.
    }
    \label{tab:transfer_matrix}
    \resizebox{0.95\textwidth}{!}{%
    \begin{tabular}{l|c|c|c|c}
        \toprule
        & \multicolumn{4}{c}{\textbf{Target MLLM (Evaluation)}} \\
        \textbf{Source Policy (Training)} & \textbf{Gemma 3 4B} & \textbf{Qwen 2.5 7B} & \textbf{Phi-3.5-vision} & \textbf{InternVL2-8B} \\
        \midrule
        \textit{Baseline: kNN} & \textit{7.27} & \textit{6.54} & \textit{5.96} & \textit{7.38} \\
        \midrule
        \textbf{LSD (Gemma)}    & \cellcolor{gray!20}6.27 & \textbf{5.58} & 6.05 & 10.62 \\
        \textbf{LSD (Qwen)}     & 6.81 & \cellcolor{gray!20}\textbf{5.95} & 6.17 & 9.69 \\
        \textbf{LSD (InternVL)} & 6.01 & 5.77 & \textbf{5.06} & \cellcolor{gray!20}8.87 \\
        \bottomrule
    \end{tabular}%
    }
\end{table*}

\subsection{Detailed Transfer Performance Plots}
\begin{figure*}[p]
    \centering
    \textbf{\large Source Policy: Gemma 3 4B-it} \par \medskip
    \begin{subfigure}[b]{0.24\linewidth}
        \centering
        \includegraphics[width=\linewidth]{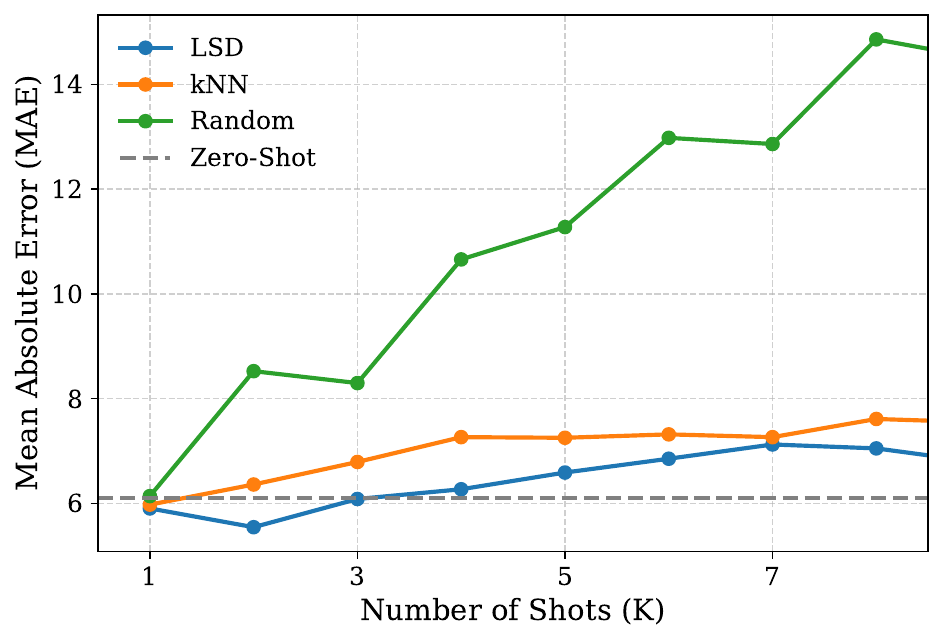}
        \caption{Target: Gemma (Intra)}
        \label{fig:supp_source_gemma_target_gemma}
    \end{subfigure}
    \hfill
    \begin{subfigure}[b]{0.24\linewidth}
        \centering
        \includegraphics[width=\linewidth]{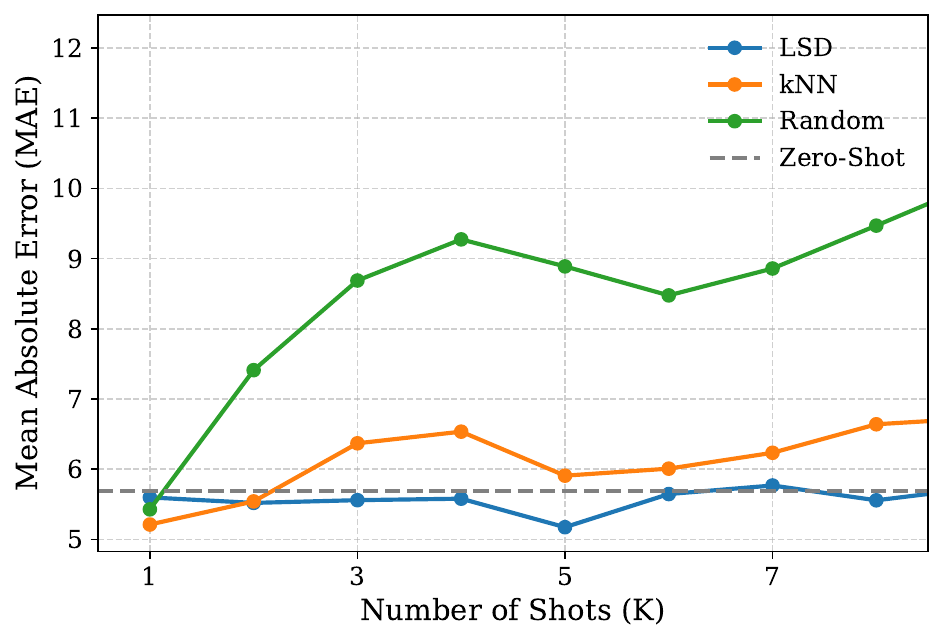}
        \caption{Target: Qwen 2.5}
        \label{fig:supp_source_gemma_target_qwen}
    \end{subfigure}
    \hfill
    \begin{subfigure}[b]{0.24\linewidth}
        \centering
        \includegraphics[width=\linewidth]{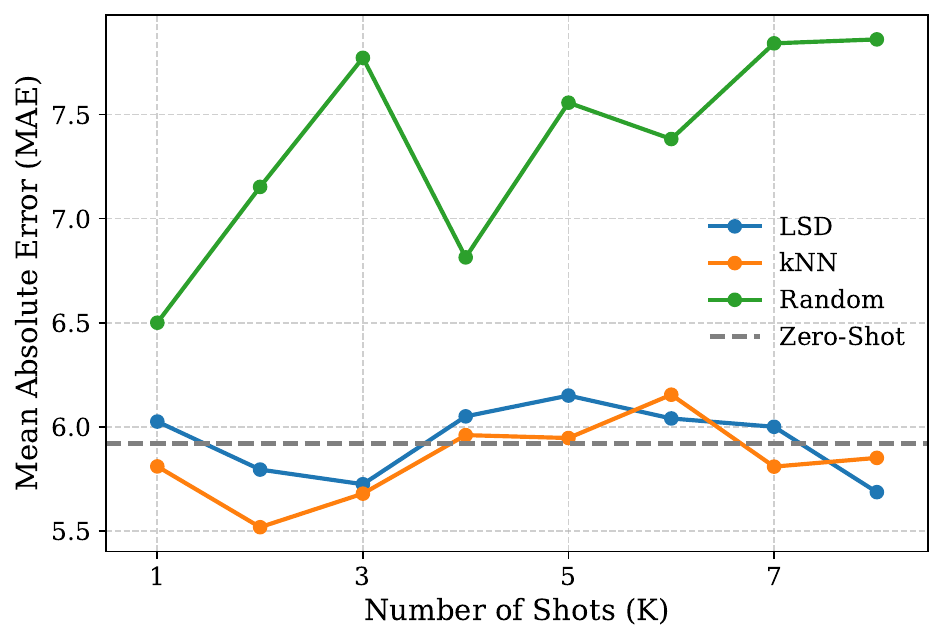}
        \caption{Target: Phi-3.5}
        \label{fig:supp_source_gemma_target_phi}
    \end{subfigure}
    \hfill
    \begin{subfigure}[b]{0.24\linewidth}
        \centering
        \includegraphics[width=\linewidth]{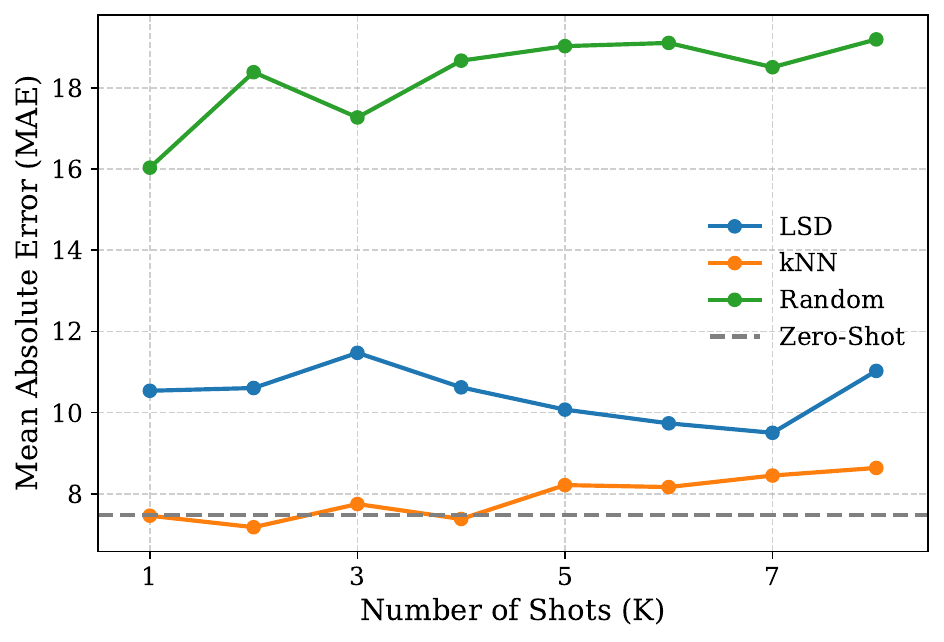}
        \caption{Target: InternVL2-8B}
        \label{fig:supp_source_gemma_target_internvl}
    \end{subfigure}
    \caption{
        \textbf{Transfer Scaling for Source Policy: Gemma 3 4B-it.}
        Performance of the Gemma-trained LSD policy evaluated across all four target models. The policy generalizes well, consistently beating kNN on Qwen and InternVL, and matching it on Phi.
    }
    \label{fig:supp_source_gemma}
\end{figure*}

\begin{figure*}[p]
    \centering
    \textbf{\large Source Policy: Qwen 2.5 7B} \par \medskip
    \begin{subfigure}[b]{0.24\linewidth}
        \centering
        \includegraphics[width=\linewidth]{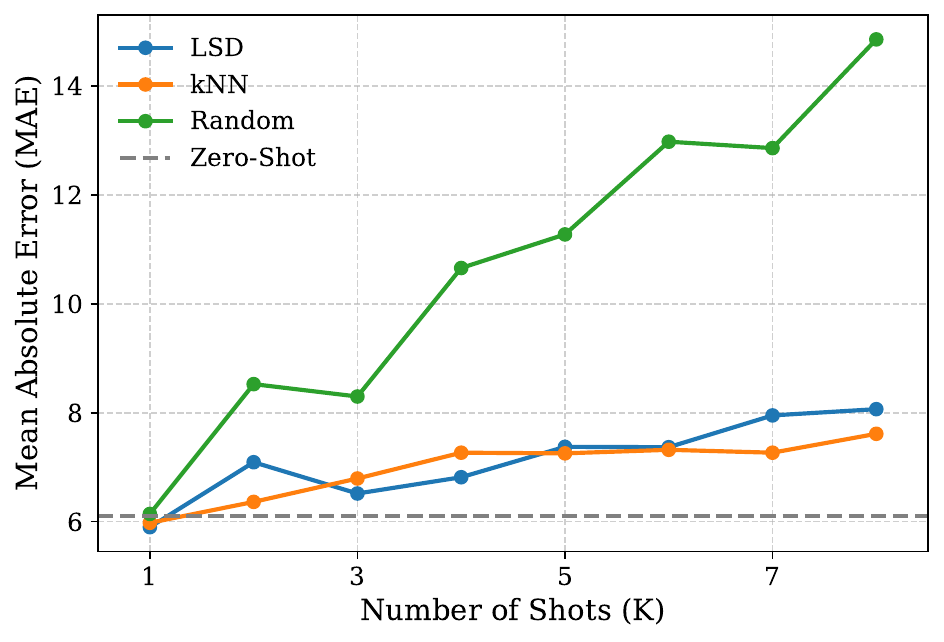}
        \caption{Target: Gemma 3}
    \end{subfigure}
    \hfill
    \begin{subfigure}[b]{0.24\linewidth}
        \centering
        \includegraphics[width=\linewidth]{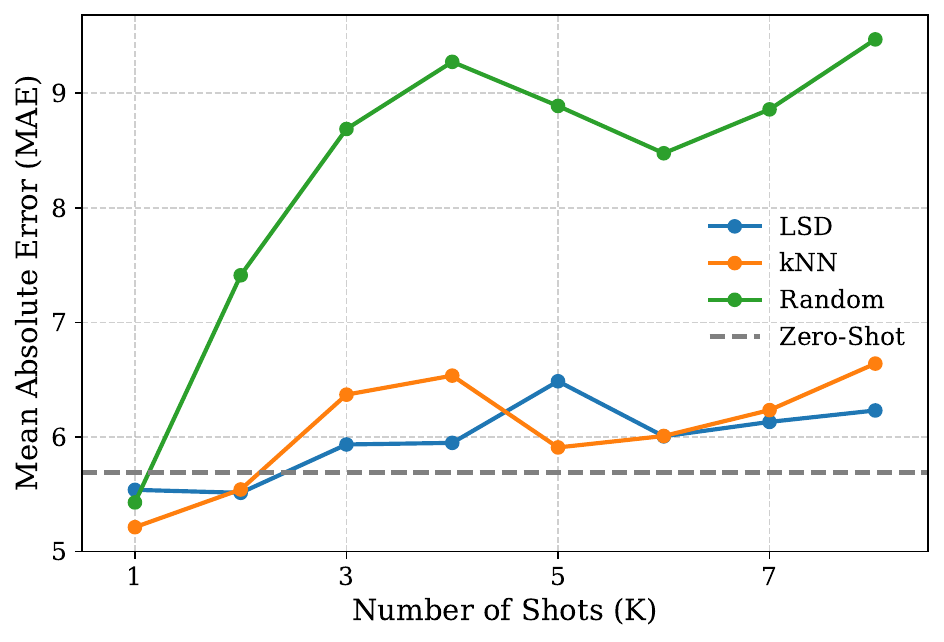}
        \caption{Target: Qwen (Intra)}
    \end{subfigure}
    \hfill
    \begin{subfigure}[b]{0.24\linewidth}
        \centering
        \includegraphics[width=\linewidth]{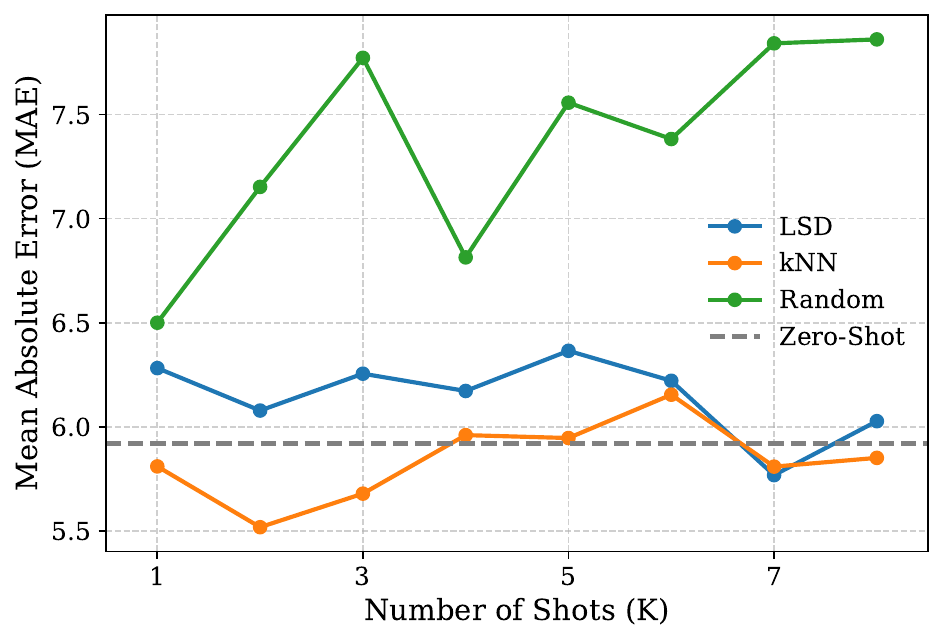}
        \caption{Target: Phi-3.5}
    \end{subfigure}
    \hfill
    \begin{subfigure}[b]{0.24\linewidth}
        \centering
        \includegraphics[width=\linewidth]{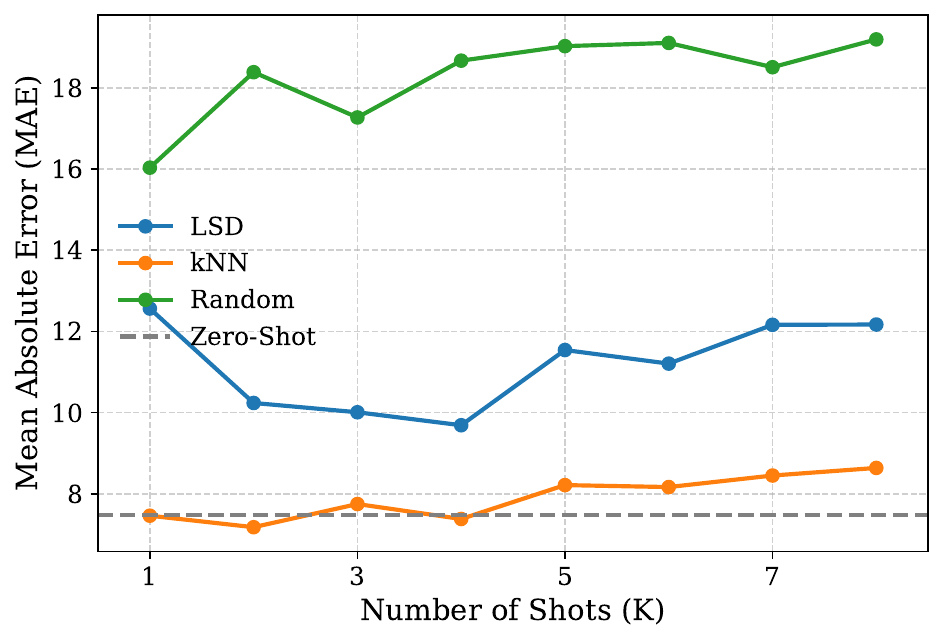}
        \caption{Target: InternVL2-8B}
    \end{subfigure}
    \caption{
        \textbf{Transfer Scaling for Source Policy: Qwen 2.5 7B.} Performance of the Qwen-trained LSD policy evaluated across all targets.
    }
    \label{fig:supp_source_qwen}
\end{figure*}

\begin{figure*}[p]
    \centering
    \textbf{\large Source Policy: InternVL2-8B} \par \medskip
    \begin{subfigure}[b]{0.24\linewidth}
        \centering
        \includegraphics[width=\linewidth]{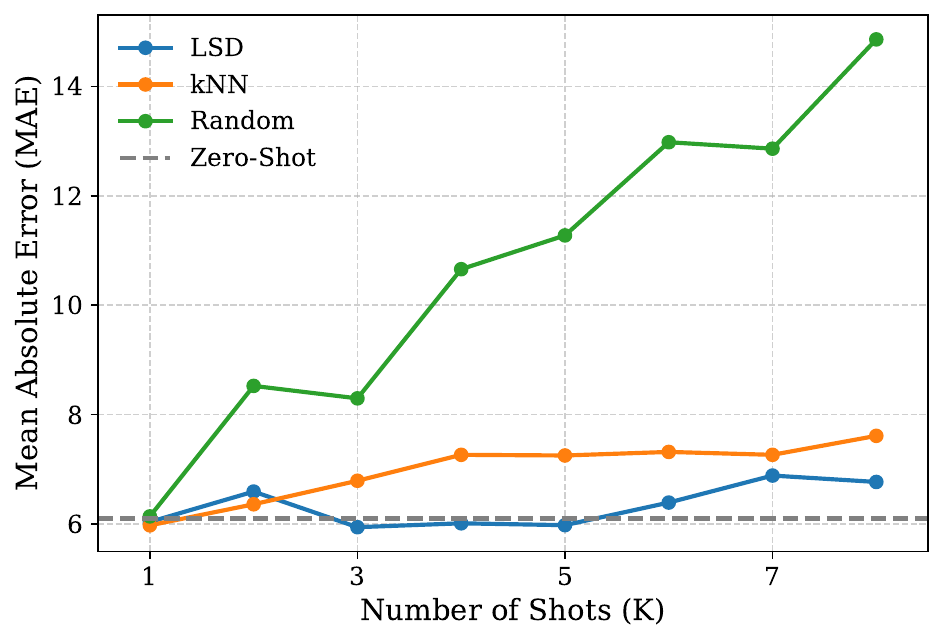}
        \caption{Target: Gemma 3}
    \end{subfigure}
    \hfill
    \begin{subfigure}[b]{0.24\linewidth}
        \centering
        \includegraphics[width=\linewidth]{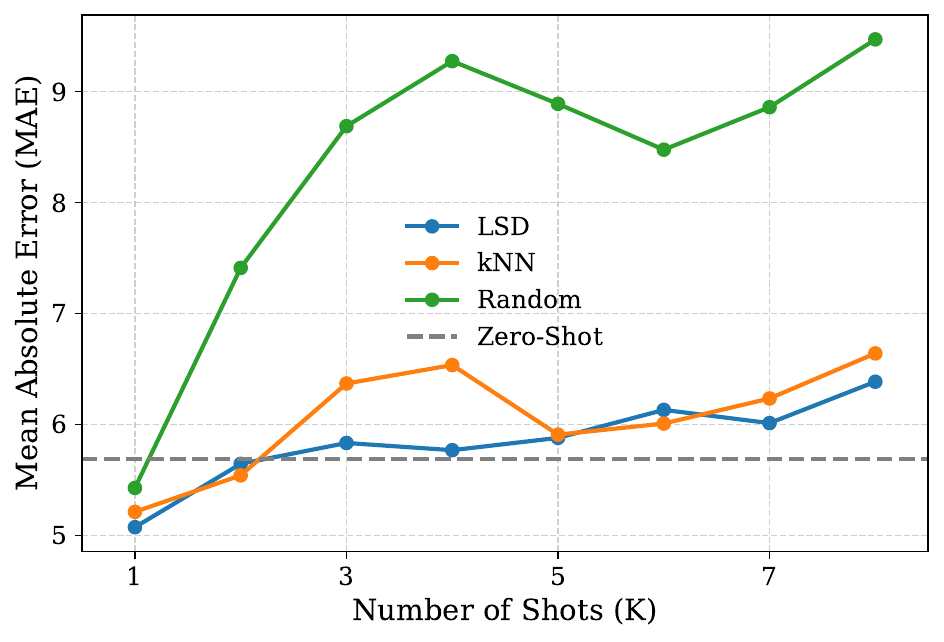}
        \caption{Target: Qwen 2.5}
    \end{subfigure}
    \hfill
    \begin{subfigure}[b]{0.24\linewidth}
        \centering
        \includegraphics[width=\linewidth]{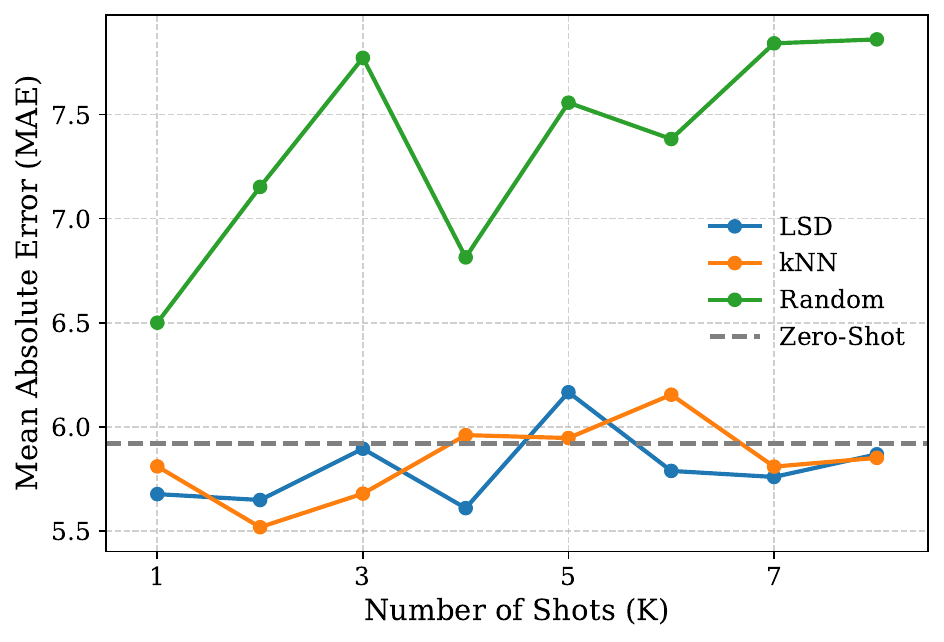}
        \caption{Target: Phi-3.5}
    \end{subfigure}
    \hfill
    \begin{subfigure}[b]{0.24\linewidth}
        \centering
        \includegraphics[width=\linewidth]{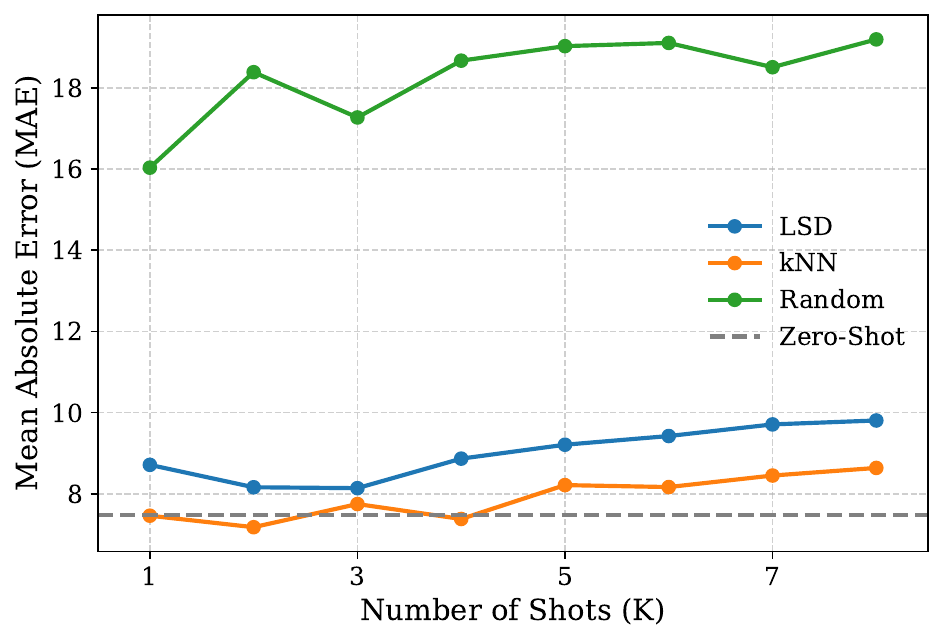}
        \caption{Target: InternVL (Intra)}
    \end{subfigure}
    \caption{
        \textbf{Transfer Scaling for Source Policy: InternVL2-8B.} Performance of the InternVL-trained LSD policy evaluated across all targets.
    }
    \label{fig:supp_source_internvl}
\end{figure*}

To visualize the robustness of these policies as the number of demonstrations ($K$) increases, we plot the performance curves for every combination of Source and Target models in Figures \ref{fig:supp_source_gemma}, \ref{fig:supp_source_qwen}, and \ref{fig:supp_source_internvl}.


\subsection{Analysis of Transfer Patterns}

This comprehensive analysis yields three critical insights into the transferability of learned retrieval policies:

\begin{itemize}
    \item \textbf{Superiority over Random Baselines:} Across all transfer scenarios (both Intra and Cross), the LSD policy consistently and significantly outperforms random selection (green lines in Figures). This confirms that the agent learns a fundamental, valid retrieval heuristic—likely selecting diverse anchors—that is universally more effective than chance, regardless of the target model.

    \item \textbf{The Specialization Gap (LSD vs. kNN):}
    \begin{itemize}
        \item \textbf{Intra-Model (Diagonal):} When the source and target models match (e.g., Gemma $\to$ Gemma), LSD consistently outperforms the strong kNN baseline. This indicates the agent learns to exploit model-specific sensitivities to specific examples or ordering.
        \item \textbf{Cross-Model (Off-Diagonal):} When transferring to a new model (e.g., Gemma $\to$ Qwen), the performance gap narrows. LSD performs comparably to kNN, sometimes slightly better or worse depending on the specific pairing. This suggests that while the ``diversity'' heuristic is universal, the fine-grained ``optimality'' of a specific set is highly coupled to the inference dynamics of the specific MLLM used during training.
    \end{itemize}

    \item \textbf{Model-Specific Sensitivity:} We observe that \textbf{Phi-3.5-vision} appears particularly resistant to policy transfer, with cross-trained policies often converging to kNN-level performance but rarely exceeding it. This highlights that different MLLM architectures (e.g., Phi vs. Qwen) may rely on fundamentally different internal mechanisms for in-context learning, limiting the direct transferability of a specialized policy.
\end{itemize}

\section{Cross-Dataset Generalization Analysis}
\label{sec:cross_dataset}

In the main paper, we presented the performance of agents trained specifically for their target domains (\textbf{LSD-Self}). To evaluate the universality of the learned retrieval policy, we employ a \textbf{LSD-Cross} protocol: we take the agent trained solely on \textbf{UTKFace} (Age Prediction) and evaluate it directly on the remaining datasets without fine-tuning. 

\subsection{Results and Discussion}
\begin{figure*}[t!]
    \centering
    \begin{subfigure}[b]{0.48\linewidth}
        \centering
        \includegraphics[width=\linewidth]{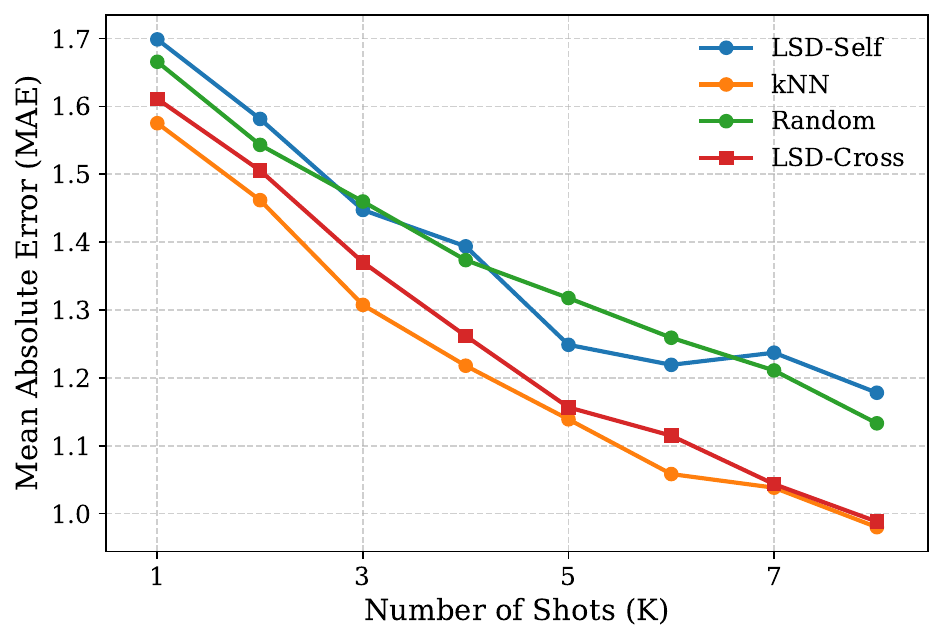}
        \caption{AVA (Aesthetic Rating)}
        \label{fig:cross_ava}
    \end{subfigure}
    \hfill
    \begin{subfigure}[b]{0.48\linewidth}
        \centering
        \includegraphics[width=\linewidth]{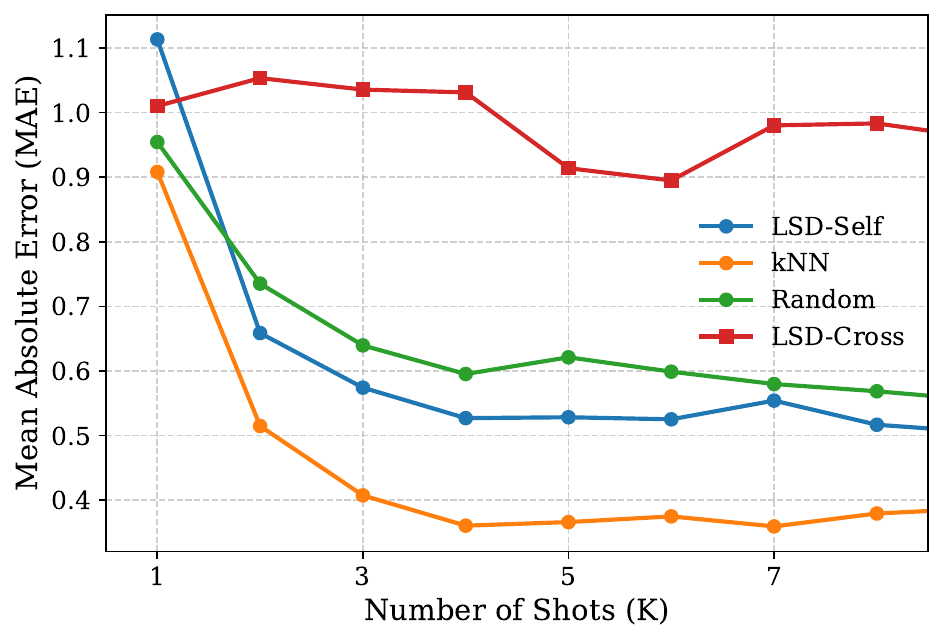}
        \caption{SCUT-FBP5500 (Facial Beauty)}
        \label{fig:cross_scut}
    \end{subfigure}
    
    \vspace{0.2cm}
    \begin{subfigure}[b]{0.48\linewidth}
        \centering
        \includegraphics[width=\linewidth]{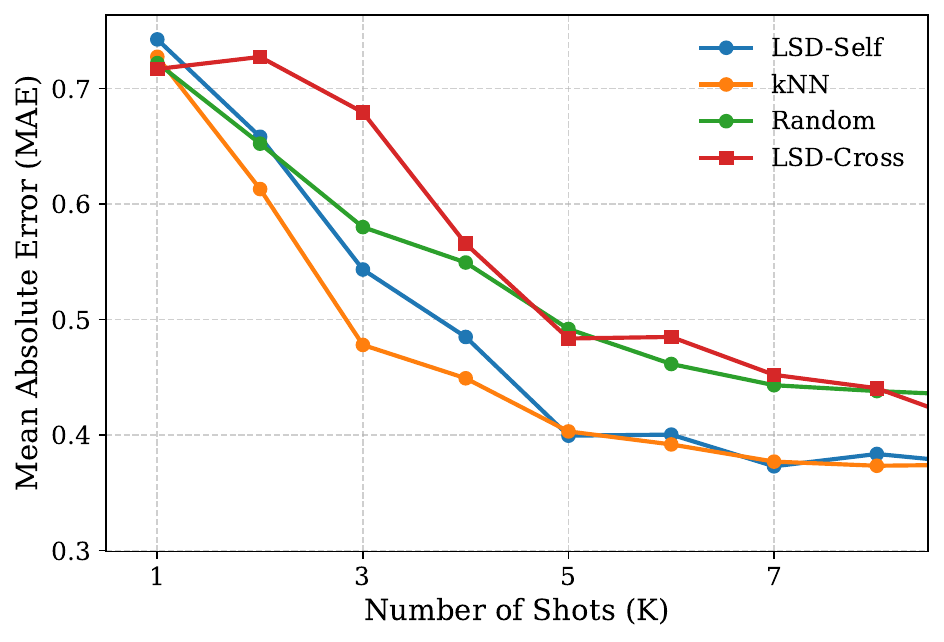}
        \caption{KonIQ-10k (Wild Image Quality)}
        \label{fig:cross_koniq}
    \end{subfigure}
    \hfill
    \begin{subfigure}[b]{0.48\linewidth}
        \centering
        \includegraphics[width=\linewidth]{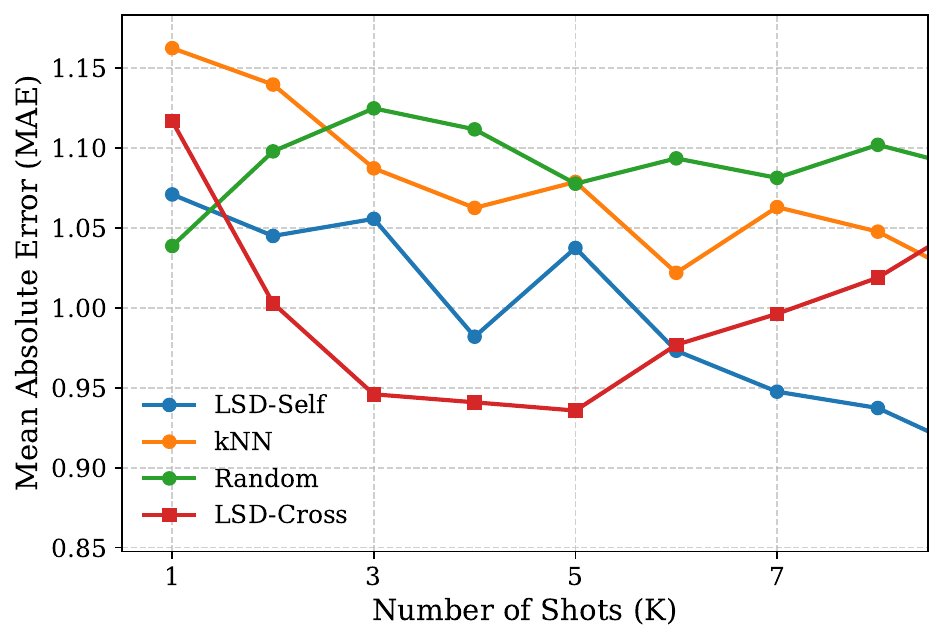}
        \caption{KADID-10k (Distorted Image Quality)}
        \label{fig:cross_kadid}
    \end{subfigure}
    
    \caption{
        \textbf{Cross-Dataset Generalization Results.} 
        We compare \textbf{LSD-Self} (Blue, trained on target), \textbf{LSD-Cross} (Red, trained on Age), \textbf{kNN} (Orange), and \textbf{Random} (Green).
        \textbf{(d) Successful Transfer:} On KADID-10k, the Age policy (Red) transfers remarkably well, matching the Self-trained agent and beating kNN/Random. 
        \textbf{(b) Negative Transfer:} On SCUT-FBP5500, the Age policy hurts performance, performing worse than random.
        \textbf{(a) Generic Policy:} On AVA, the Cross and Self policies perform identically, suggesting the learned retrieval strategy is task-agnostic but inferior to semantic matching (kNN) for aesthetics.
    }
    \label{fig:cross_plots}
\end{figure*}

The results are visualized in \cref{fig:cross_plots}. The transfer performance varies significantly by task nature, revealing distinct behaviors of the learned policy.

\textbf{Robust Transfer on Objective Distortions (KADID-10k).} 
As shown in \cref{fig:cross_kadid}, the \textbf{LSD-Cross} agent (Red) demonstrates exceptional transfer capabilities on the KADID-10k dataset. Despite being trained on faces, the policy—which learns to select diverse anchors to span the regression range—is highly effective for Image Quality Assessment. It matches the performance of the domain-specific \textbf{LSD-Self} agent and significantly outperforms the kNN baseline. This confirms our hypothesis that for objective regression tasks, a ``diversity-aware'' selection strategy is universally beneficial and task-agnostic.

\textbf{Negative Transfer on Facial Analysis (SCUT-FBP5500).} 
In \cref{fig:cross_scut}, we observe a case of negative transfer. The Age-trained policy performs significantly worse than Random selection. We hypothesize this is due to conflicting objectives: the UTKFace agent is incentivized to retrieve a maximally diverse age range (e.g., toddlers and the elderly). However, for facial beauty scoring, extreme age diversity may introduce noise or out-of-distribution examples that confuse the MLLM's attractiveness estimation.

\textbf{Generic Heuristics on Aesthetics (AVA).} 
On the AVA dataset (\cref{fig:cross_ava}), the \textbf{LSD-Cross} (Red) and \textbf{LSD-Self} (Blue) lines are nearly indistinguishable. This implies that training specifically on AVA yielded the same generic retrieval strategy as training on Age. However, both fall short of the kNN baseline. This reinforces the finding that for subjective, content-heavy tasks like aesthetics, semantic similarity (kNN) remains the dominant factor, and the ``diversity'' heuristic learned by LSD provides less benefit.

\begin{table*}[h!]
    \centering
    \caption{
        \textbf{Cross-Dataset Generalization Analysis.}
        We report the MAE ($\downarrow$) for the \textbf{kNN} baseline, the Domain-Specific Agent (\textbf{Self}), and the Cross-Trained Agent (\textbf{Cross}, trained on UTKFace) across $K \in \{1, 4, 8\}$.
        \textbf{LSD-Self} typically sets the upper bound.
        Comparing \textbf{LSD-Cross} to \textbf{kNN} reveals where the learned ``diversity'' heuristic transfers effectively (e.g., KADID-10k) versus where domain-specific visual matching is superior (e.g., AVA).
    }
    \label{tab:cross_results_full}
    \resizebox{\textwidth}{!}{%
    \begin{tabular}{l|c|ccc|ccc|ccc}
        \toprule
        & & \multicolumn{3}{c|}{\textbf{K=1}} & \multicolumn{3}{c|}{\textbf{K=4}} & \multicolumn{3}{c}{\textbf{K=8}} \\
        \cmidrule(lr){3-5} \cmidrule(lr){6-8} \cmidrule(lr){9-11}
        \textbf{Dataset} & \textbf{0-Shot}
            & \textbf{kNN} & \textbf{Self} & \textbf{Cross}
            & \textbf{kNN} & \textbf{Self} & \textbf{Cross}
            & \textbf{kNN} & \textbf{Self} & \textbf{Cross} \\
        \midrule
        AVA & 1.38
            & \textbf{1.58} & 1.70 & 1.61
            & \textbf{1.22} & 1.26 & 1.39
            & \textbf{0.98} & 0.99 & 1.18 \\
        SCUT-FBP5500 & 1.07
            & \textbf{0.91} & 1.13 & 1.01
            & \textbf{0.36} & 0.53 & 1.03
            & \textbf{0.38} & 0.52 & 0.98 \\
        KonIQ-10k & 0.78
            & \textbf{0.73} & 0.74 & 0.72
            & \textbf{0.45} & 0.48 & 0.57
            & \textbf{0.37} & 0.38 & 0.44 \\
        KADID-10k & 1.13
            & 1.16 & \textbf{1.07} & 1.12
            & 1.06 & \textbf{0.98} & 0.94
            & 1.05 & \textbf{0.94} & 1.02 \\
        \bottomrule
    \end{tabular}
    }
\end{table*}

\section{Extended Qualitative Analysis}
\begin{figure*}[t]
    \centering
    \includegraphics[width=\linewidth]{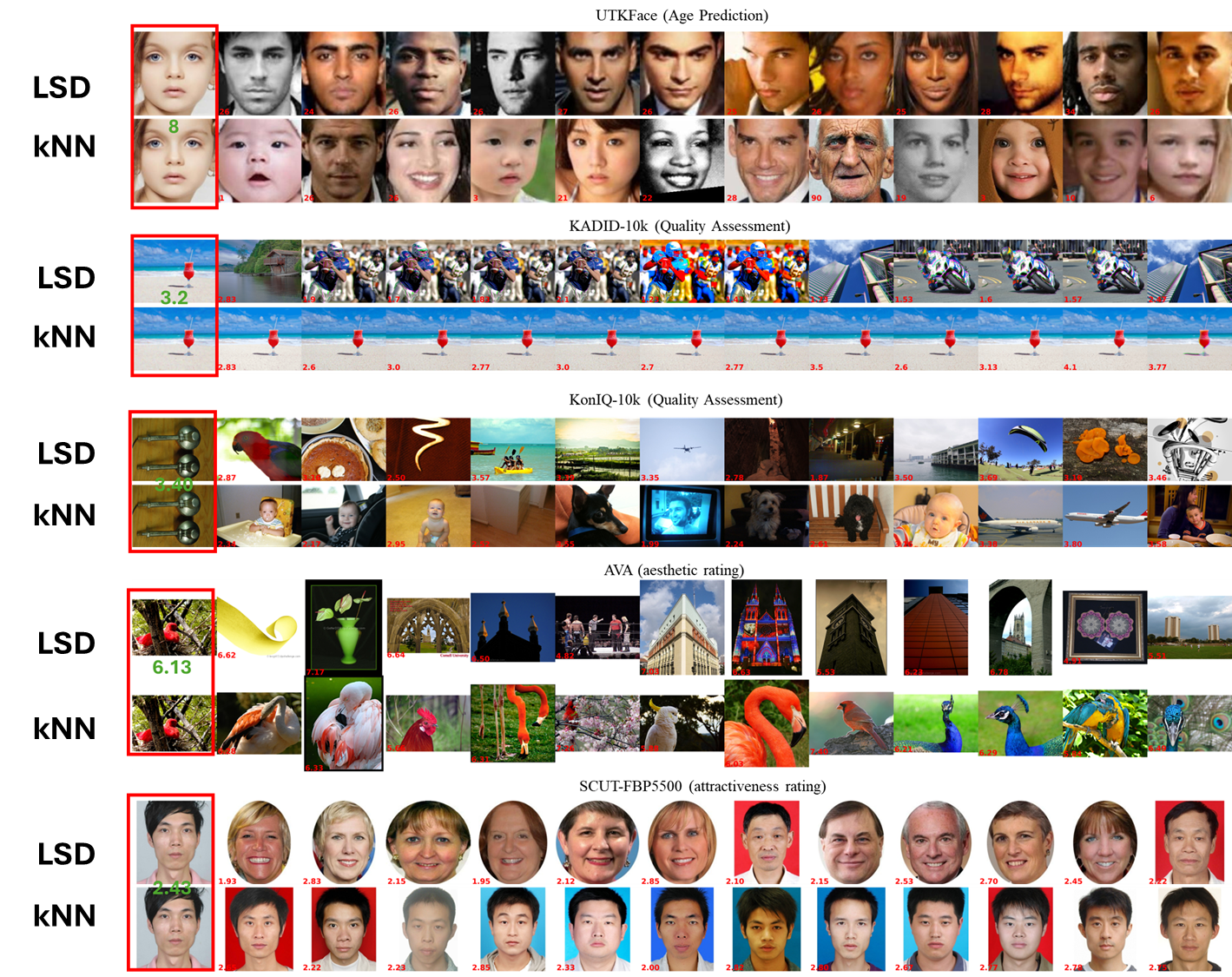}
    \caption{
        \textbf{Extended Qualitative Comparison of Selected Demonstrations ($K=11$) across benchmark datasets.}
        \textbf{Row 1: UTKFace (Age).} For a query of a \textbf{young child}, the \textbf{kNN} baseline retrieves a homogeneous set of other children and babies. In contrast, \textbf{LSD (Ours)} retrieves a diverse timeline of faces, ranging from toddlers to adults and the elderly, providing the VLM with a complete regression scale.
        \textbf{Row 2: KADID-10k (Quality).} For a query of a beach scene, \textbf{kNN} fails by retrieving near-duplicate versions of the \emph{same source image}, adding zero new information. \textbf{LSD} selects visually distinct scenes (sports, cityscapes) with varying distortion types.
        \textbf{Row 3: KonIQ-10k (Quality).} For a query of abstract metal spheres, \textbf{LSD} retrieves a broad semantic range (animals, food, landscapes), whereas kNN gets stuck in a narrow cluster of indoor/portrait shots.
        \textbf{Row 4: AVA (Aesthetics).} For a query of red berries, \textbf{kNN} relies on color and content matching, retrieving flamingos and other birds. \textbf{LSD} ignores the specific content, selecting architecture and objects to illustrate broad aesthetic principles.
        \textbf{Row 5: SCUT-FBP5500 (Beauty).} For a query of an Asian male, \textbf{kNN} exhibits high demographic bias, retrieving only other Asian males. \textbf{LSD} retrieves a diverse set of demographics (varying gender and race), reducing bias and helping the model score attractiveness independent of demographic features.
    }
    \label{fig:qualitative_supp}
\end{figure*}

In \cref{fig:qualitative_supp}, we visualize the retrieval behavior of the proposed LSD agent versus the kNN baseline. The results highlight a critical limitation of standard retrieval in regression tasks: \emph{semantic redundancy}.

\paragraph{Overcoming Semantic Redundancy (KADID-10k \& AVA).}
The most distinct failure mode of kNN is visible in the KADID-10k example (Row 2). Because the dataset contains multiple distorted versions of the same reference images, kNN retrieves 11 versions of the \emph{same beach scene}. This provides the VLM with no comparative information regarding quality standards. LSD, driven by the reward signal, learns to avoid this redundancy, selecting completely different scenes (sports, traffic, buildings) to illustrate the concept of ``image quality'' broadly. Similarly, in AVA (Row 4), kNN matches the red color of the query berries to flamingos, whereas LSD retrieves structurally diverse images (architecture, objects) that likely span the aesthetic scoring range.

\paragraph{Demographic and Age Diversity (SCUT-FBP5500 \& UTKFace).}
For facial analysis tasks, kNN tends to over-index on demographic similarity.
\begin{itemize}
    \item On \textbf{UTKFace} (Row 1), the kNN baseline retrieves almost exclusively babies and toddlers for a child query. This prevents the VLM from accessing ``anchor'' examples of adults or the elderly, which are necessary to calibrate the upper bounds of age estimation. LSD retrieves a full age spectrum.
    \item On \textbf{SCUT-FBP5500} (Row 5), kNN restricts the context to the same gender and ethnicity (Asian males) as the query. LSD breaks this demographic lock, retrieving Caucasian and Asian faces of both genders, which encourages the VLM to abstract the concept of "facial beauty" away from specific demographic features.
\end{itemize}

\paragraph{Subjective Tasks (AVA, SCUT-FBP5500).}
For the subjective tasks, the behavioral difference remains distinct, even if the quantitative advantage is smaller.
\begin{itemize}
    \item On \textbf{AVA} (Fig. \ref{fig:qualitative_supp}b), kNN retrieves images with near-identical composition (e.g., 12 sunsets), effectively asking the model to ``rate this sunset based on these other sunsets.'' LSD retrieves a diverse portfolio of photography styles (e.g., macro, portrait, landscape). While kNN performed better quantitatively on AVA, LSD's policy is demonstrably more informative about the \emph{general concept} of aesthetics, rather than just specific object aesthetics.
    \item Similarly, on \textbf{SCUT-FBP5500} (Fig. \ref{fig:qualitative_supp}c), kNN selects faces that look like ``siblings'' of the query. LSD selects a cohort that varies significantly in appearance and attractiveness rating, attempting to provide a broader comparative scale for the MLLM.
\end{itemize}

\end{document}